\documentclass[letterpaper]{article} 
\PassOptionsToPackage{table}{xcolor}
\usepackage{aaai2027}
\nocopyright
\usepackage{algorithm}
\usepackage{algorithmic}  
\usepackage[hyphens]{url}  
\usepackage{graphicx} 
\usepackage{natbib}  
\usepackage{caption} 
\usepackage{amsmath}
\usepackage{amsfonts}
\usepackage{booktabs}
\usepackage{multirow}
\title{NAS-PO: Native Attention-Strategy Policy Optimization for Vision-Language Models}
\author{
Ke An\textsuperscript{\rm 1}\equalcontrib, Shizhe Xiang\textsuperscript{\rm 2}\equalcontrib, Yue Liu\textsuperscript{\rm 3}, Liang Chen\textsuperscript{\rm 1},\\
Pei Fu\textsuperscript{\rm 4}, Shaojie Zhang\textsuperscript{\rm 4}, Xiaoxing Hu\textsuperscript{\rm 5}, Yupei Wang\textsuperscript{\rm 1}\corresponding
}
\affiliations{
\textsuperscript{\rm 1}Beijing Institute of Technology\\
\textsuperscript{\rm 2}Tianjin University\\
\textsuperscript{\rm 3}University of Chinese Academy of Sciences\\
\textsuperscript{\rm 4}Xiaomi Inc.\\
\textsuperscript{\rm 5}Shanghai Jiao Tong University
}

\newcommand{\AppTOCEntry}[4]{%
  \addtocontents{toc}{\protect\contentsline{#1}%
    {\protect\numberline{#2}#4}{\thepage}{}}%
}
\newcommand{\AppSection}[2]{%
  \section{#2}%
  \AppTOCEntry{section}{\thesection}{#1}{#2}%
}
\newcommand{\AppSubsection}[2]{%
  \subsection{#2}%
  \AppTOCEntry{subsection}{\thesubsection}{#1}{#2}%
}

\begin{document}

\maketitle

\begin{abstract}
Reinforcement learning with verifiable rewards (RLVR) has become an effective post-training paradigm for vision-language models (VLMs). Standard RLVR, however, assigns identical outcome-derived credit to all generated tokens regardless of their perceptual relevance. Recent perception-aware methods estimate visual reliance through image perturbation or masking. These indirect proxies require perturbation branches, while corruption sensitivity does not necessarily identify the visual evidence used for reasoning. We instead shape model perception through native attention, an intrinsic model signal. Rather than treating attention as a post-hoc explanation, we formulate it as a reward-anchored, token-aligned routing signal that encourages attention patterns supporting visually grounded reasoning.
To this end, we propose \textbf{NAS-PO}, a \textbf{N}ative
\textbf{A}ttention-\textbf{S}trategy \textbf{P}olicy \textbf{O}ptimization framework. NAS-PO analyzes native attention with a visual mass policy that quantifies visual allocation and a conditional visual policy that captures spatial routing over visual tokens, making the optimization well-posed. Based on this,
Native Attention-Strategy Advantage Shaping (NAS-AS) draws on the visual-mass policy to redistribute signed policy-gradient credit for visually engaged rollouts. The Adaptive Attention-Coupled (AAC) loss then aligns the visual-mass policy on all response tokens and the conditional visual policy on visually active ones, consolidating the visual-routing patterns that drive successful trajectories. Experiments demonstrate that NAS-PO substantially surpasses leading RL-tuned VLMs across 7 out-of-domain reasoning and perception benchmarks, with its effectiveness consistently validated across 8B and 32B. Visual-Routing Sensitivity (VRS) analysis further indicates that the optimized attention prioritizes visually load-bearing regions.
\end{abstract}


\section{Introduction}

Reinforcement learning with verifiable rewards (RLVR) has become a leading post-training
paradigm for large language models, with Group Relative Policy Optimization (GRPO)
replacing costly step-level supervision with group-relative outcome
feedback~\citep{shao2024deepseekmath}. Extending this paradigm to vision-language models
(VLMs) is well motivated because many multimodal reasoning tasks admit automatically
verifiable answers. Standard RLVR, however, assigns identical outcome-derived credit to
all tokens within a generated response, irrespective of their perceptual roles.
Consequently, the objective identifies successful trajectories without specifying
whether visual evidence contributed to generation or where relevant evidence was
selected.

Recent perception-aware methods estimate visual reliance by comparing model outputs
under original and perturbed or masked
images~\citep{wang2025perception,huang2025spotlight,rezaei2026cppo}. These approaches
establish visual dependence as a useful signal for multimodal policy optimization. Their
corruption-sensitivity estimates nevertheless remain indirect because an output change
need not identify the visual evidence used in the original reasoning process. They also
require auxiliary perturbation branches. These limitations motivate deriving a
fine-grained visual-routing signal directly from the model's behavior on clean rollouts.

\begin{figure}[t]
\centering
\includegraphics[width=1\columnwidth]{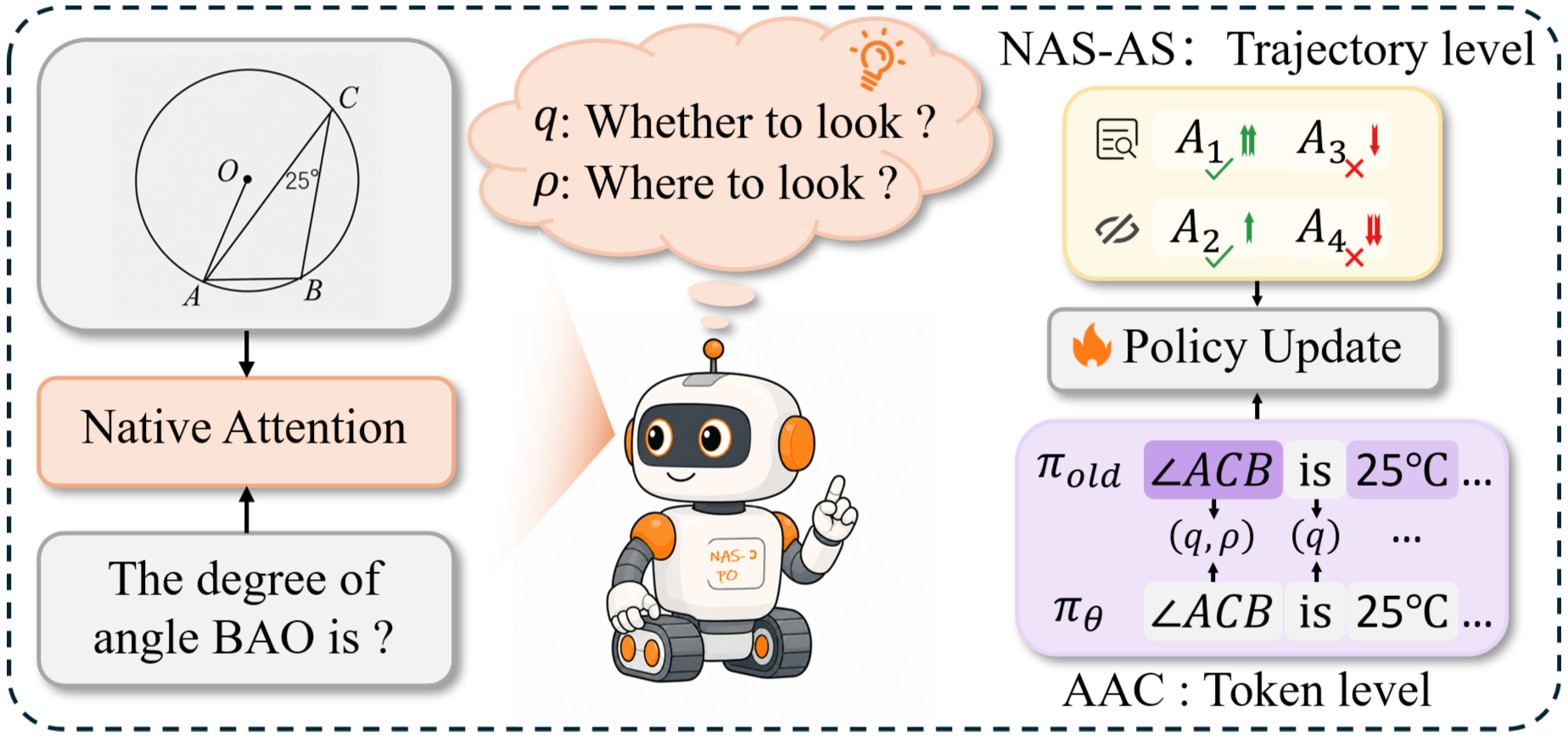} 
  \caption{NAS-PO decomposes the native attention into two policies: visual-mass policy $q$ for whether to rely on the image, and conditional visual policy $\rho$ for where to attend. Native Attention-Strategy Advantage Shaping (NAS-AS) optimizes this routing at trajectory level and the Adaptive Attention-Coupled (AAC) loss at token level.}
\label{fig1}
\end{figure}

Native attention provides such a signal. During autoregressive generation, it specifies
how each response token allocates attention over visual and textual context and remains
differentiable with respect to the policy parameters~\citep{attention}. We do not interpret a
last-layer attention map as a complete causal account of model
behavior~\citep{jain2019attention,wiegreffe2019attention}. Instead, we formulate native
attention as a reward-anchored, token-aligned routing signal. The verifier identifies
successful trajectories, while native attention characterizes the model's allocation of
visual evidence within them. Their combination enables outcome feedback to optimize
visual routing on clean rollouts.

Native attention must be structured before it can serve as an optimization objective. A
raw attention map couples two distinct decisions: the allocation of attention between
visual and non-visual context, and the distribution of visual attention across image
regions. As shown in Fig.~\ref{fig1}, we losslessly decompose the visual component into a
visual-mass policy and a conditional visual policy. The former quantifies whether and to
what extent a response token relies on the image, whereas the latter captures the spatial
distribution of attention conditional on visual engagement. Their composition exactly
recovers the original visual attention, making visual reliance and spatial routing
separately optimizable at their appropriate scopes.

Building on this decomposition, we propose \textbf{NAS-PO}, a Native Attention-Strategy
Policy Optimization framework that operates at two levels. At the trajectory level,
Native Attention-Strategy Advantage Shaping (NAS-AS) summarizes visual engagement through
the visual-mass policy and redistributes signed policy-gradient credit across rollouts
while preserving the overall update scale. This allows trajectory updates to reflect
differences in visual reliance without replacing the verifier reward. NAS-AS therefore
places greater learning emphasis on visually engaged rollouts, reinforcing them more
strongly when the answer is correct and suppressing them more strongly when it is
incorrect. At the token level,
the Adaptive Attention-Coupled (AAC) loss consolidates routing patterns associated with
positive-advantage trajectories. It aligns the visual-mass policy across all valid
response tokens and the conditional visual policy only where visual attention is active.
Failed trajectories continue to contribute through the standard policy-gradient
objective but are not treated as negative attention targets because an incorrect answer
does not localize the failure to perception. NAS-PO therefore operates on clean rollouts
without additional perturbation branches or verifier calls.

We conduct extensive experiments across seven challenging out-of-domain multimodal
reasoning and perception benchmarks spanning mathematical, geometric, logical, and
multidisciplinary reasoning. Using Qwen3-VL models, NAS-PO achieves relative average
gains of 7.08\% and 7.21\% over GRPO at 8B and 32B, respectively. Component ablations
confirm the complementary roles of NAS-AS and AAC, while checkpoint-wise evaluations
show that the gains persist throughout training. Finally, by contrasting the effects of
removing highly and weakly attended visual regions, Visual-Routing Sensitivity (VRS)
provides intervention-based evidence that NAS-PO more effectively prioritizes visually
load-bearing regions~\citep{fong2017meaningful,serrano2019attention}.

In summary, our main contributions are as follows:

\begin{itemize}
    \item We formulate native attention as a reward-anchored, token-aligned routing signal
    and losslessly decompose it into visual-mass and conditional visual policies, enabling
    direct optimization of visual reliance and spatial routing without auxiliary
    perturbation proxies.
    \item We propose NAS-PO, combining NAS-AS to redistribute trajectory credit according
    to visual engagement while preserving the update scale, and AAC to consolidate
    token-level routing on positive-advantage trajectories without treating failed
    trajectories as negative attention targets.
    \item Across seven out-of-domain benchmarks, NAS-PO achieves relative average gains of
    7.08\% and 7.21\% over GRPO at 8B and 32B, respectively. Ablations, training dynamics,
    and VRS further validate the complementarity of its components, the persistence of its
    gains, and the visual relevance of the learned routing.

\end{itemize}

\section{Related Work}
\paragraph{Reinforcement Learning with Verifiable Reward.} RLVR has become a powerful post-training paradigm for improving the reasoning abilities of large language models. GRPO~\citep{shao2024deepseekmath} derives group-relative advantages from sparse verifiable feedback over sampled responses, replacing the critic-based advantage estimation commonly used in PPO-style RLHF. Building on this group-based RLVR paradigm, a series of optimizers have been proposed to improve training stability, efficiency, and scalability, including DAPO~\citep{yu2026dapo}, GSPO~\citep{zheng2025group}, and GMPO~\citep{zhao2025geometric}. Recent variants further refine GRPO-style training from different perspectives, such as correcting length bias, introducing value-based augmentation, or encouraging more diverse and stable reward utilization~\citep{liu2025understanding, yue2025vapo, chen2025dra, dong2025agentic}. Yet in VLMs, outcome-only rewards cannot tell visual grounding from language-prior shortcuts, nor localize failures to perception or reasoning~\citep{li2023evaluating, guan2024hallusionbench, tong2024eyes}. 
This gap motivates perception-aware multimodal RL strategies that explicitly account for the model's use of visual evidence.

\paragraph{Perception-Aware Strategies in Multimodal RL.} Recent multimodal RL methods address the visual blind spot of outcome-only RLVR by introducing perception-aware optimization signals~\citep{huang2025vision}. Some methods design perception-verifiable or spatially grounded rewards for visual tasks and grounded reasoning~\citep{liu2025visual,shen2025satori, ni2026point}. Others estimate visual reliance through image perturbations or counterfactual inputs, and reshapes policy updates accordingly~\citep{wang2025perception, huang2025spotlight, rezaei2026cppo}. Complementary efforts improve visual grounding through structured perception-reasoning roles or privileged visual supervision, such as crop-conditioned self-distillation and answer-free tutoring signals~\citep{miao2026seeing,yuan2026vision,xiang2026teaching}. These methods demonstrate the importance of perception-aware training for MLLMs, but they typically rely on external visual interventions, predefined response roles, or privileged contexts.

\section{Method}

\subsection{Preliminary: Group Relative Policy Optimization (GRPO)}
\label{sec:grpo}

Given a multimodal prompt $x = (I, \mathcal{Q})$ comprising an image $I$ and a textual query $\mathcal{Q}$, the rollout policy $\pi_{\mathrm{old}}$ samples a group of $G$ responses $\{y_i\}_{i=1}^{G}$. In RLVR framework, each full response $y_i = (y_{i,1}, \dots, y_{i,T_i})$ receives a trajectory-level binary reward $R_i \in \{0, 1\}$ based solely on whether its final extracted answer
matches the ground truth. GRPO converts the group rewards into normalized advantages:
\begin{equation}
A_i=\frac{R_i-\mathrm{mean}(R_{1:G})}{\mathrm{std}(R_{1:G})+\epsilon}.
\end{equation}
For token $y_{i,t}$, the probability ratio between the current and old policies is
\begin{equation}
r_{i,t}(\theta)=\frac{\pi_\theta(y_{i,t}\mid x,y_{i,<t})}{\pi_{\mathrm{old}}(y_{i,t}\mid x,y_{i,<t})},
\end{equation}
and the policy is updated to maximize the standard clipped objective, where the uniform advantage $A_i$ is broadcast to every valid response token~\citep{shao2024deepseekmath,schulman2017ppo}:

\begin{equation}
\begin{split}
\mathcal{J}_{\text{GRPO}}(\theta) = \mathbb{E} \bigg[ \frac{1}{G} \sum_{i=1}^{G} \frac{1}{T_i} \sum_{t} M_{i,t} \min \Big( r_{i,t}(\theta)A_i, \\
\operatorname{clip}(r_{i,t}(\theta), 1-\varepsilon, 1+\varepsilon) A_i \Big) \bigg],
\end{split}
\end{equation}

with $M_{i,t}$ the response-token mask and $T_i=\sum_t M_{i,t}$. While the objective is scalable, its credit assignment is \textit{text-centric}. Under a single binary reward, the same scalar $A_i$ is broadcast to every token of $y_i$. As a result, GRPO leaves two internal decisions of multimodal generation entirely unsupervised~\citep{wang2025perception,huang2025spotlight,rezaei2026cppo}:

\begin{itemize}
    \item \textit{\textbf{Whether}} a generated token relied on visual evidence 
    \item \textit{\textbf{Where}} the model attended to when visual evidence was needed. 

\end{itemize}

These two unsupervised dimensions of visual routing are precisely the targets that NAS-PO addresses next.

\subsection{Native Attention-Strategy Decomposition}
\label{sec:naspo}

\begin{figure*}[!t]
\centering
\includegraphics[width=1\linewidth]{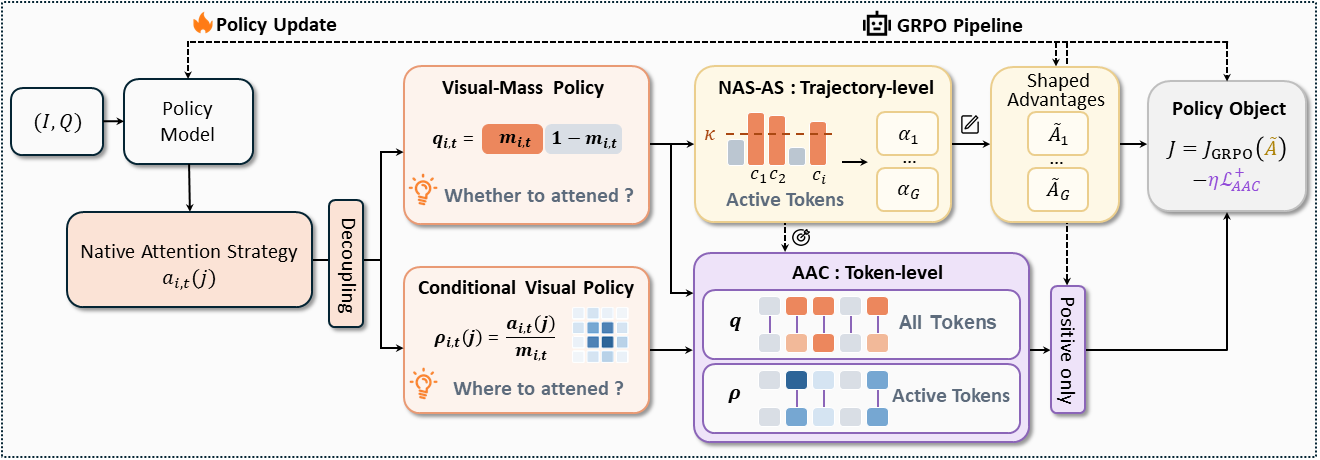}
\caption{Overview of NAS-PO. Given an image--query pair $(I,\mathcal{Q})$, the standard GRPO pipeline produces $G$ clean rollouts and verifier-derived group-relative advantages $A_i$. Using the same clean rollouts, NAS-PO reads native attention $a_{i,t}(j)$ and losslessly decomposes its visual component into a visual-mass policy $q_{i,t}$, which captures whether and how much a response token relies on the image, and a conditional visual policy $\rho_{i,t}$, which captures where attention is routed over visual tokens. The visual-mass policy also provides visually active indicators $c_{i,t}$ shared by NAS-AS and the conditional branch of AAC. At the trajectory level, NAS-AS summarizes visual reliance into a bounded factor $\alpha_i$ and forms $\tilde{A}_i=\alpha_iA_i$. At the token level, AAC aligns $q$ on all valid response tokens and $\rho$ on visually active ones under the unshaped positive-advantage gate $[A_i]_+$. The shaped policy-gradient objective and the adaptively weighted AAC loss jointly update the policy.}
\label{fig2}
\end{figure*}

Under sparse trajectory-level rewards, native attention is an intrinsic and differentiable
signal that records how each generated token routes information from the visual context.
It is available on GRPO's clean rollouts and remains differentiable during the actor update,
so it can be read as a fixed behavioral statistic from $\pi_{\mathrm{old}}$ and optimized as
a trainable object in the actor. We therefore treat it not as a post-hoc interpretability
artifact but as a \textbf{native attention-strategy}, a visual-routing optimization object
that we extract, decompose, and couple back into the RLVR loop~\citep{attention}.

\textbf{Native Attention Extraction.} Let $\mathcal{V}_i$ be the visual-token indices of
sample $i$, and let $P_i$ be the serialized prompt length. Let $p_{i,t}=P_i+t-1$
be the query position that scores $y_{i,t}$, and let
$\mathcal{C}_{i,t}=\{j:j\le p_{i,t}\}$ denote its causal support. We extract the
head-averaged attention from the last decoder layer as
\begin{equation}
a_{i,t}(j)
=\frac{1}{H}\sum_{h=1}^{H}
\operatorname{Softmax}_{k\in\mathcal{C}_{i,t}}\!\left(
\frac{Q^{(h)}_{i,p_{i,t}}K^{(h)\top}_{i,k}}{\sqrt{d_k}}
\right)_{\!j}.
\end{equation}

where $H$ is the number of heads and $Q^{(h)},K^{(h)}$ are the post-RoPE queries and keys
from the last decoder layer~\citep{su2024roformer,qwen2025qwen3vl}. Using the same query that scores $y_{i,t}$ keeps the attention
summary aligned with the token-level policy update and avoids an off-by-one or pre-RoPE
attention proxy. During training, detached old Q/K provide fixed targets, while differentiable
current Q/K let the attention objective update the model through its routing logits. This gives
NAS-PO a token-aligned, differentiable visual-routing signal without external image perturbations.
Appendix~A.1 and Appendix~B.2 detail the extraction, decomposition, post-RoPE reconstruction,
and response-token alignment.

\textbf{Attention Strategy Decomposition.} The visual component of native attention $a_{i,t}(j)$, $j\in\mathcal{V}_i$, combines two complementary aspects of routing: the global allocation of attention to visual tokens and its spatial distribution within the image. To separate their optimization roles, NAS-PO factorizes this visual component into its total mass and a conditional distribution:
\begin{equation}
\label{eq3}
a_{i,t}(j) = m_{i,t} \cdot \rho_{i,t}(j), \quad j \in \mathcal{V}_i,
\end{equation}
where $m_{i,t} = \sum_{j \in \mathcal{V}_i} a_{i,t}(j) \in [0, 1]$ is the total visual mass. This exact factorization provides the basis for the two components of our Native Attention Strategy:

\begin{itemize}

\item \textbf{Visual-Mass Policy $q_{i,t}$ (The \textit{Whether} Decision):} To represent how much attention a response token allocates to the image, we define the visual-mass policy
\begin{equation}
q_{i,t} = [\,m_{i,t},\, 1-m_{i,t}\,],
\end{equation}
This two-category marginal summarizes the head-averaged attention assigned to visual and nonvisual causal positions. It is defined for every valid response token and provides the visual-allocation signal used by NAS-AS and AAC.

\item \textbf{Conditional Visual Policy $\rho_{i,t}$ (The \textit{Where} Decision):} Conditioned on nonzero visual mass, the conditional visual policy describes how the visual component is routed over visual tokens:
\begin{equation}
\rho_{i,t}(j) = \frac{a_{i,t}(j)}{m_{i,t}}, \quad j \in \mathcal{V}_i.
\end{equation}
For $m_{i,t}>0$, $\rho_{i,t}$ is a proper distribution over the visual context. The identity $a_{i,t}(j)=m_{i,t}\rho_{i,t}(j)$ therefore exactly reconstructs the visual component of the head-averaged attention map.
\end{itemize}

This decomposition separates visual allocation from within-image routing while preserving their conditional structure. The visual-mass policy $q$ is defined on every valid response token, whereas the conditional visual policy $\rho$ is meaningful only when nontrivial visual mass is present. AAC can therefore align $q$ broadly while applying $\rho$ alignment only on the active visual support. The two policies share the same native-attention pathway but provide distinct optimization targets and gating conditions.

\subsection{Trajectory-Level Credit Reallocation with NAS-AS}
\label{sec:nas_as}

A binary reward cannot tell a correct answer reasoned through the image from one reached
by a text-only shortcut, so both receive the same advantage. NAS-AS mitigates this
ambiguity by reweighting each trajectory's advantage using the old visual-mass policy
$q^{\mathrm{old}}$ as a detached, per-trajectory summary of visual engagement.

\textbf{Visual Reliance Summary.} Many text-centric tokens (e.g., function words) inherently lack visual evidence, so aggregating every token would inject perceptual noise. To filter it, NAS-AS calibrates a per-batch adaptive threshold $\kappa$ on detached visual mass:
\begin{equation}
\kappa = \mathrm{Quantile}_{\delta}\!\left(\{\,m_{i,t}^{\mathrm{old}} : M_{i,t}=1\,\}\right).
\end{equation}
We set $\delta=0.25$ to exclude the lowest-mass quartile of valid tokens from the relative active support. Appendix~B.3 provides the distributed implementation and a sensitivity analysis. The resulting per-token visually active indicator
\begin{equation}
\label{eq9}
c_{i,t} = M_{i,t}\,\mathbb{I}\!\left[\,m_{i,t}^{\mathrm{old}} > \kappa\,\right]
\end{equation}
defines the relative active support shared by NAS-AS and the conditional branch of the AAC loss.

Once $\kappa$ partitions tokens into active and inactive sets, we summarize rollout-level visual reliance by active-token coverage and average excess visual mass. The coverage
\begin{equation}
b_i \;=\; \frac{1}{T_i}\sum_{t} c_{i,t}
\end{equation}
measures the fraction of response tokens in the relative active support. The average excess mass
\begin{equation}
u_i \;=\; \frac{\sum_t c_{i,t}\,\sqrt{m_{i,t}^{\mathrm{old}} - \kappa\,}}{\sum_t c_{i,t} + \epsilon}
\end{equation}
measures the visual mass above $\kappa$ within this support. The square root reduces the influence of isolated high-mass tokens (Appendix~A.2). NAS-AS combines them as
\begin{equation}
s_i \;=\; b_i\,u_i,
\end{equation}
which is high only when visual engagement is both frequent and substantial.

\textbf{Trajectory Advantage Reshape.} To transfer the visual-reliance score into a credit-assignment signal, $s_i$ is normalized to $z_i \in [0, 1]$ and mapped to a bounded per-trajectory scaling factor:
\begin{equation}
\alpha_i = \beta_{\min} + z_i(\beta_{\max} - \beta_{\min}).
\end{equation}
Here, $\beta_{\min}=0.9$, and $\beta_{\max}$ is adaptively calibrated for each nondegenerate update micro-batch $\mathcal U$ such that $\frac{1}{|\mathcal U|}\sum_{i\in\mathcal U}\alpha_i\approx1$ (Appendix~A.2). The shaped advantage is
\begin{equation}
\tilde{A}_i = \alpha_i A_i.
\end{equation}
Because $\alpha_i>0$, NAS-AS preserves the verifier-determined direction while assigning larger-magnitude positive or negative updates to more visually engaged rollouts, concentrating the calibrated update budget on image-conditioned behavior.

The shaped advantage $\tilde{A}_i$ enters only the probability surrogate, whereas AAC gates on the unshaped $[A_i]_+$. Coupling both objectives through $\alpha_i$ would grant a high-visual-mass trajectory a larger policy update and a stronger alignment signal simultaneously, forming a self-reinforcing loop that inflates attention regardless of task relevance (Appendix~A.2).

\subsection{Token-Level Visual-Routing Consolidation with AAC}
\label{sec:aac}

At the token level, AAC complements attention-space policy objectives that optimize native
attention distributions during RL~\citep{li2026reinforcedattention}. NAS-AS determines how strongly each rollout influences the policy update but does not specify how visual evidence should be allocated and routed at individual response tokens. AAC therefore treats the native routing patterns of positive-advantage rollouts as reward-anchored references. Aligning the current strategy $(q^\theta,\rho^\theta)$ with the detached reference $(q^{\mathrm{old}},\rho^{\mathrm{old}})$ converts sparse success feedback into fine-grained guidance for consolidating visual allocation and spatial routing.


\textbf{Decoupled Attention-Strategy Divergence.} AAC aligns the visual-mass policy $q$ and the conditional visual policy $\rho$ at their respective scopes:
\begin{equation}
D_{i,t}=\underbrace{\mathrm{JSD}\!\left(q^\theta_{i,t}\,\|\,q^{\mathrm{old}}_{i,t}\right)}_{D_q,\ \text{all valid tokens}}
+\;c^\rho_{i,t}\,\underbrace{\mathrm{JSD}\!\left(\rho^\theta_{i,t}\,\|\,\rho^{\mathrm{old}}_{i,t}\right)}_{D_\rho,\ \text{conditional support}},
\end{equation}
where $c^\rho_{i,t}$ denotes the support on which both the reference and current strategies allocate nontrivial visual mass. The visual-mass policy is aligned on every valid response token because the decision of whether to use visual information is always defined. Conditional spatial routing is aligned only after visual mass has been allocated, avoiding an artificial spatial target for tokens that do not rely on the image.

We use JSD because it is symmetric and remains finite when the current and reference supports differ, yielding a well-defined alignment objective even when their attention distributions have different sharpness~\citep{lin1991divergence}. Numerical stabilization is detailed in Appendix~A.3.

\textbf{Positive-Anchored Aggregation.} A binary failure does not reveal whether the error originates from visual routing or from later reasoning, computation, or formatting. AAC therefore does not treat failed rollouts as negative routing targets. Instead, it consolidates routing patterns associated with positive-advantage trajectories:
\begin{equation}
L^{+}_{\mathrm{AAC}}=
\mathbb{E}_{(i,t)\sim\mathcal{U}_{\mathrm{tok}}}
\!\left[[A_i]_+\,\mathrm{sg}(w^+_{i,t})\,D_{i,t}\right],
\end{equation}
where $\mathcal{U}_{\mathrm{tok}}$ contains the valid response tokens in the current update micro-batch, $[A_i]_+=\max(A_i,0)$, and $w^+_{i,t}$ is the clipped importance weight. Positive advantage anchors the alignment to verified success, while importance weighting keeps it consistent with the policy update. Failed rollouts still contribute through the probability surrogate but are not used as attention counterexamples, consolidating success-associated routing without mislabeling later reasoning errors as attention errors. With detached references and weights, AAC acts directly on the current native-attention pathway rather than as an output-level auxiliary objective (Appendix~A.3).

\begin{table*}[htbp]
\centering
\renewcommand{\arraystretch}{1.2}
{\small\setlength{\tabcolsep}{1.2pt}%
\begin{tabular}{l|ccccc|c|ccc|c|c}
\toprule
\multirow{2}{*}{\textbf{Model}} & \multicolumn{6}{c}{\textbf{General Multimodal Reasoning}} & \multicolumn{4}{c|}{\textbf{Vision-Dependent Multimodal Reasoning}} & \multirow{2}{*}{\textbf{Overall}} \\
\cmidrule(lr){2-7} \cmidrule(lr){8-11}
& \textbf{MMK12} & \textbf{Geo3k} & \textbf{MathVerse} & \textbf{MathVista} & \textbf{We-Math} & \textbf{AVG} & \textbf{MMMU-Pro} & \textbf{MathVerse$_V$} & \textbf{LogicVista} & \textbf{AVG} & \\
\midrule
OneThinker-8B & 47.19 & 41.90 & 64.3 & 69.47 & 49.36 & 54.44 & 40.75 & 43.54 & 48.60 & 44.30  & 50.64 \\
NoisyRollout-8B & 68.56 & 56.23 & 72.17 & 68.45 & 77.02 & 68.49  & 40.76 & 69.73 & 52.39 & 54.29  & 63.16 \\
PAPO-8B & 68.85 & 55.39 & 73.96 & 68.95 & 77.21 & 68.87 & 42.42 & 70.25 & 54.11 & 55.59  & 63.89 \\
VPPO-8B & 75.30 & 58.98 & 76.98 & 73.08 & 75.51 & 71.97 & 41.78 & 74.68 & 52.43 & 56.30  & 66.09 \\
CPPO-8B & 68.90 & 58.38 & 75.55 & 69.26 & 77.67 & 69.95 & 42.39 & 71.95 & 53.1 & 55.81  & 64.65 \\
\midrule
Qwen3-VL-8B-Instruct & 26.28 & 43.57 & 30.75 & 63.30 & 46.73 & 42.13 & 19.03 & 29.38 & 39.01 & 29.14 & 37.26 \\
+ DAPO & 70.44 & 56.71 & 75.13 & 73.45 & 76.81 & 70.51  & 40.75 & 71.99 & 52.06 & 54.93  & 64.67 \\
\rowcolor{gray!15} + DAPO \& NAS-PO & 76.73 & 61.31 & 78.85 & 74.77 & 78.15 & 73.96  & 43.29 & 75.48 & 54.91 & 57.89  & 67.94 \\
+ GSPO & 68.91 & 57.77 & 75.46 & 74.10 & 76.38 & 70.52 & 41.63 & 71.75 & 51.45 & 54.94  & 64.68 \\
\rowcolor{gray!15} + GSPO \& NAS-PO & 75.45 & 58.02 & 78.42 & 75.15 & 75.67 & 72.54 & 42.25 & 75.18 & 54.89 & 57.44 & 66.88 \\
+ GRPO & 68.50 & 55.57 & 72.89 & 71.76 & 76.78 & 69.10 & 40.78 & 69.18 & 52.83 & 54.26 & 63.54 \\
\rowcolor{gray!15} + GRPO \& NAS-PO & 77.49 & 60.13 & 78.36 & 75.24 & 78.09 & 73.86 & 43.78 & 75.61 & 55.60 & 58.33 & 68.04 \\
\midrule
\multicolumn{12}{c}{\textit{Scaling to Larger Model}} \\
\midrule
Qwen3-VL-32B-Instruct & 41.87 & 56.84 & 37.60 & 70.96 & 58.33 & 53.12 & 22.80 & 36.51 & 48.46 & 35.92 & 46.67 \\
+ GRPO & 80.66 & 64.26 & 77.70 & 71.86 & 79.89 & 74.87 & 49.41 & 74.66 & 58.61 & 60.89 & 69.63 \\
\rowcolor{gray!15} + GRPO \& NAS-PO & 89.43 & 69.89 & 82.02 & 79.26 & 82.91 & 80.70 & 52.35 & 79.00 & 62.37 & 64.57 & 74.65 \\
\bottomrule
\end{tabular}
}
\caption{Main results (avg@8 acc $\%$) on general and more vision-dependent multimodal reasoning tasks. MathVerse$_V$ refers to the vision-centric subset of MathVerse. AVG and Overall are unweighted means over the displayed evaluation configurations.}
\label{tab:multimodal_reasoning_results}
\end{table*}

\subsection{Overall NAS-PO Objective}
\label{sec:naspo_objective}

\textbf{Adaptive Coupling.} AAC should guide visual routing without overwhelming verifier-driven policy optimization. Because its raw magnitude evolves during training, a fixed coefficient can make the alignment signal negligible or disproportionately strong. We therefore match its detached scalar magnitude to a target fraction of the policy surrogate:
\begin{equation}
\eta=\mathrm{clip}\!\left(
\tau_{\mathrm{AAC}}
\frac{\left|\mathrm{sg}\!\left(\mathcal{J}_{\mathrm{PG}}(\tilde{A})\right)\right|}
{\left|\mathrm{sg}\!\left(L_{\mathrm{AAC}}^{+}\right)\right|+\epsilon},
\eta_{\min},\eta_{\max}
\right),
\end{equation}
We set $\tau_{\mathrm{AAC}}=0.3$ and clip $\eta$ to $[\eta_{\min},\eta_{\max}]$, preventing AAC from vanishing or dominating while preserving verifier-driven policy learning as the primary objective. Appendix~A.4 provides the implementation and loss-scale analysis, and Appendix~B.5 discusses computational overhead.

Combining the Native Attention-Strategy Advantage Shaping (NAS-AS) and Adaptive Attention-Coupled (AAC) Loss with the GRPO objective yields our complete \textbf{NAS-PO objective}:

\begin{equation}
\begin{split}
    \mathcal{J}_{\mathrm{NAS\text{-}PO}}(\theta) &= \mathbb{E} \Bigg[ \frac{1}{G} \sum_{i=1}^{G} \frac{1}{T_i} \sum_t M_{i,t} \\
    &\quad \cdot \min\Big( r_{i,t}(\theta)\,\alpha_i A_i, \\
    &\quad \quad \mathrm{clip}(r_{i,t}(\theta), 1-\varepsilon, 1+\varepsilon)\,\alpha_i A_i \Big) \Bigg] \\
    &\quad - \eta\,L_{\mathrm{AAC}}^{+}(\theta).
\end{split}
\label{eq:nas_po}
\end{equation}

\begin{figure}[!b]
\centering
\includegraphics[width=\columnwidth]{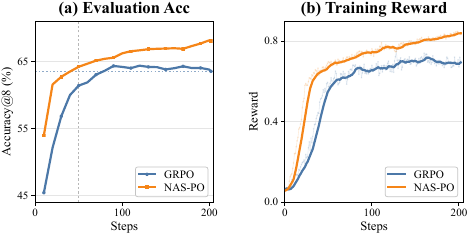}
\caption{Training dynamics on Qwen3-VL-8B. (a) Average Accuracy@8 across eight configurations from seven out-of-domain benchmarks. The dotted guides mark GRPO's final accuracy and step 50, where NAS-PO has already exceeded it. (b) Reward on ViRL39K. Faint lines show per-step values and solid lines show 10-step running averages. NAS-PO leads at all evaluation checkpoints and nearly all reward steps, showing faster learning from the early stages and a sustained advantage over GRPO.}
\label{fig:training_dynamics}
\end{figure}

\section{Experiments}

\subsection{Experimental Setup}

\subsubsection{Training Dataset.} We train our model on ViRL39K~\citep{wang2025vlrethinker}, a dataset containing 38,870 verifiable vision-language question-answering pairs. ViRL39K spans diverse domains, from grade-school problems to STEM and social topics, and includes tasks that require reasoning over charts, diagrams, tables, documents, and spatial relationships.

\subsubsection{Evaluation.} Following prior works, we use 7 out-of-domain benchmarks to evaluate our model's performance, including MMK12~\citep{meng2025mmeureka}, Geometry3K~\citep{lu2021intergps}, MathVista~\citep{lu2024mathvista}, MathVerse~\citep{zhang2024mathverse}, We-Math~\citep{qiao2025wemath}, MMMU-Pro~\citep{yue2025mmmupro}, and LogicVista~\citep{xiao2024logicvista}. These benchmarks encompass math, geometric reasoning, multi-discipline multimodal reasoning and logical reasoning tasks.  For datasets such as MathVista and MathVerse, free-form answer instances are filtered out to enable rule-based answer verification and avoid relying on LLM-as-a-judge. All evaluation metric are based on exact match against the ground-truth answer. We report average accuracy @ 8 for all benchmarks as well as the averaged results over the general, vision-dependent, and overall splits.

\subsubsection{Baselines.} We perform RL training from Qwen3-VL-8B and 32B~\citep{qwen2025qwen3vl}, comparing the standard GRPO~\citep{shao2024deepseekmath}, GSPO~\citep{zheng2025group}, and DAPO~\citep{yu2026dapo} baselines with our proposed NAS-PO variants. We also compare NAS-PO with recent RLVR methods proposed for VLMs: OneThinker~\citep{feng2025onethinker}, NoisyRollout~\citep{liu2025noisyrollout}, PAPO~\citep{wang2025perception}, VPPO~\citep{huang2025spotlight}, and CPPO~\citep{rezaei2026cppo}. OneThinker-8B uses its official model based on Qwen3-VL-8B. For the other methods, whose released models use Qwen2.5-VL-7B, we retrain their open-source implementations on Qwen3-VL-8B while strictly preserving their released training hyperparameters. For the controlled optimizer comparisons, GRPO, GSPO, DAPO, and their NAS-PO variants use the same Qwen3-VL policy models and response-length budget. Full training and NAS-PO-specific hyperparameters are summarized in Appendix~C.

\begin{table*}[t]
\centering
\renewcommand{\arraystretch}{1.2}
{\small\setlength{\tabcolsep}{0.7pt}%
\begin{tabular}{l|ccccc|c|ccc|c|cc}
\toprule
\multirow{2}{*}{\textbf{Model}}
& \multicolumn{6}{c|}{\textbf{General Multimodal Reasoning}}
& \multicolumn{4}{c|}{\textbf{Vision-Dependent Multimodal Reasoning}}
& \multicolumn{2}{c}{\textbf{Overall}} \\
\cmidrule(lr){2-7} \cmidrule(lr){8-11} \cmidrule(lr){12-13}
& \textbf{MMK12} & \textbf{Geo3k} & \textbf{MathVerse} & \textbf{MathVista} & \textbf{We-Math}
& \textbf{$\Delta_{\mathrm{rel}}^{\%}$}
& \textbf{MMMU-Pro} & \textbf{MathVerse$_V$} & \textbf{LogicVista}
& \textbf{$\Delta_{\mathrm{rel}}^{\%}$}
& \textbf{Score} & \textbf{$\Delta_{\mathrm{rel}}^{\%}$} \\
\midrule
Qwen3-VL-8B-Instruct
& 26.28 & 43.57 & 30.75 & 63.30 & 46.73 & --
& 19.03 & 29.38 & 39.01 & -- & 37.26 & -- \\
+ GRPO
& 68.50 & 55.57 & 72.89 & 71.76 & 76.78 & --
& 40.78 & 69.18 & 52.83 & -- & 63.54 & -- \\
+ GRPO \& NAS-AS
& 76.09 & 58.67 & 77.10 & 74.61 & 77.00 & $\uparrow$5.20
& 41.04 & 74.14 & 54.03 & $\uparrow$3.94 & 66.59 & $\uparrow$4.80 \\
+ GRPO \& AAC
& 77.37 & 59.94 & 77.31 & 75.18 & 77.50 & $\uparrow$6.31
& 43.10 & 74.23 & 53.77 & $\uparrow$5.10 & 67.30 & $\uparrow$5.92 \\
\rowcolor{gray!15} + GRPO \& NAS-PO
& \textbf{77.49} & \textbf{60.13} & \textbf{78.36} & \textbf{75.24} & \textbf{78.09} & $\uparrow$6.89
& \textbf{43.78} & \textbf{75.61} & \textbf{55.60} & $\uparrow$7.49 & \textbf{68.04} & $\uparrow$7.08 \\
\bottomrule
\end{tabular}}
\caption{Detailed ablation of Native Attention-Strategy Advantage Shaping (NAS-AS) and Adaptive Attention-Coupled (AAC) loss. Each component improves GRPO on every evaluation configuration, and their combination yields the best result throughout. $\Delta_{\mathrm{rel}}$ denotes the relative gain over GRPO for the corresponding aggregate.}
\label{tab:ablation_results}
\end{table*}

\begin{figure*}[!t]
\centering
\includegraphics[width=1\linewidth]{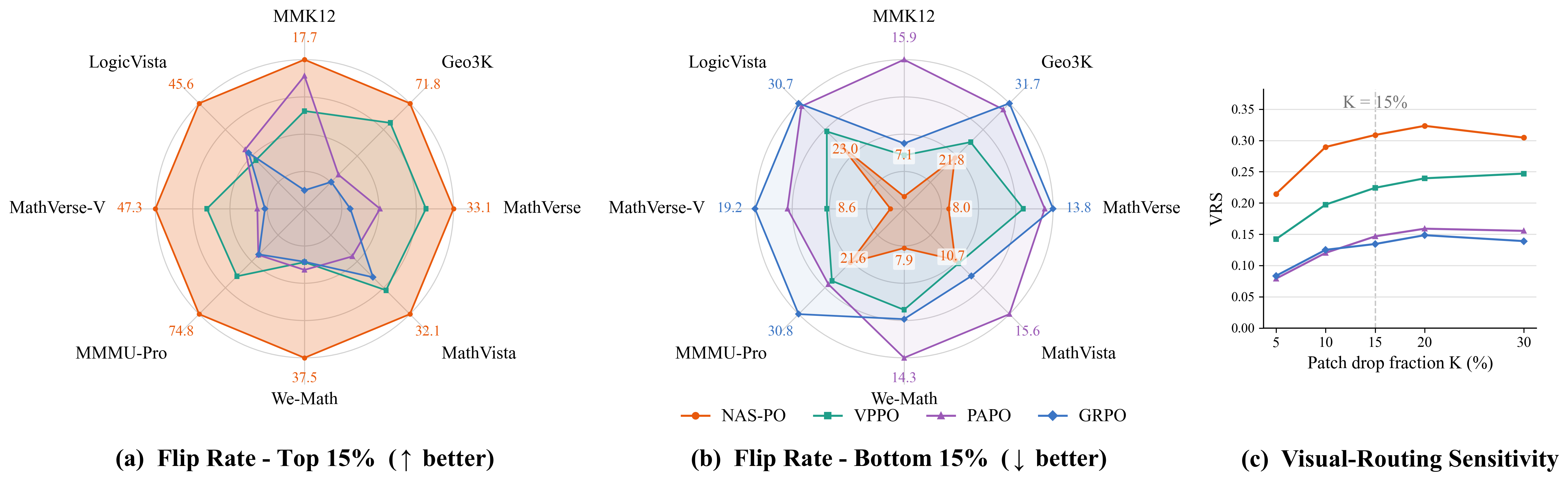}
\caption{Visual-Routing Sensitivity. (a) Dropping the Top-15\% patches flips NAS-PO far more often than the baselines on all seven benchmarks ($\uparrow$ better). (b) Dropping the Bottom-15\% barely affects it ($\downarrow$ better). (c) Their difference, VRS, remains highest for NAS-PO across every drop fraction $K$, reaching $0.31$ at $K{=}15\%$ versus $0.13$ for GRPO. Together, the interventions show that NAS-PO's attention ranking identifies visually load-bearing regions.}
\label{fig3}
\end{figure*}

\subsection{Main Results}

Table~\ref{tab:multimodal_reasoning_results} shows that GRPO with NAS-PO achieves the best overall Accuracy@8 among the evaluated 8B models, reaching 68.04, 1.95 points above VPPO-8B, while raising the vision-dependent average from 54.26 to 58.33. The gains are broad: GRPO with NAS-PO leads on MMK12, MathVista, MMMU-Pro, MathVerse$_V$, and LogicVista, while DAPO with NAS-PO leads on Geo3K, MathVerse, and We-Math. NAS-PO improves all three tested optimizers, with overall gains of 3.27, 2.20, and 4.50 points over DAPO, GSPO, and GRPO, respectively. At 32B, it raises GRPO from 69.63 to 74.65, showing that the gain persists at the larger tested model scale.

\subsection{Training Dynamics}

Figure~\ref{fig:training_dynamics} complements the final results in Table~\ref{tab:multimodal_reasoning_results} with the full training trajectory. NAS-PO leads GRPO at all 21 evaluation checkpoints and exceeds GRPO's final Accuracy@8 by step 50, 30 steps before GRPO first reaches that level. GRPO subsequently fluctuates around 64 and ends at 63.54, whereas NAS-PO continues improving to 68.04. Its ViRL39K reward also rises faster from the early stages and is higher at 199 of 202 steps. The consistent lead in both checkpoint accuracy and training reward supports a sustained optimization advantage rather than a gain confined to the final checkpoint. Appendix~D provides the full-resolution reward and per-benchmark curves. These are descriptive single-run trajectories.

\subsection{Ablation Study}

Table~\ref{tab:ablation_results} shows that each component is effective when applied individually. NAS-AS raises GRPO from 63.54 to 66.59, while AAC reaches 67.30 and improves the vision-dependent average from 54.26 to 57.03. Their combination performs best at 68.04 overall. Both components improve every individual benchmark over GRPO, and the complete NAS-PO is strongest on all eight evaluation configurations. These results indicate that trajectory-level credit shaping and token-level visual-routing regularization provide complementary gains.

\subsection{Visual-Routing Sensitivity}

Higher accuracy alone cannot tell whether NAS-PO grounds its answers in the image or merely exploits language priors. Following intervention-based faithfulness tests~\citep{fong2017meaningful,serrano2019attention}, \textbf{Visual-Routing Sensitivity (VRS)} ranks patches by the conditional visual policy $\rho$, separately removes the top- or bottom-$K\%$, and measures the difference in answer-flip rates. Figure~\ref{fig3} shows that NAS-PO is highly sensitive to its most-attended regions but nearly invariant to its least-attended ones: at $K{=}15\%$, VRS reaches $0.31$, about $2.3\times$ GRPO's $0.13$, and remains highest at every drop fraction. Because this perturbation is used only for evaluation, it provides intervention-based evidence that NAS-PO prioritizes visually load-bearing regions, without claiming a complete causal account of the model's reasoning~\citep{jain2019attention,wiegreffe2019attention}. Appendix~E gives the full protocol and tests.

\section{Conclusion}

We presented NAS-PO, a Native Attention-Strategy Policy Optimization framework for vision-language RLVR. It decomposes native attention into a visual-mass policy for whether a token uses visual evidence and a conditional visual policy for where it attends. NAS-AS reallocates trajectory-level credit by visual engagement, while AAC consolidates successful token-level routing without treating failed trajectories as attention counterexamples. NAS-PO consistently improves multiple RLVR backbones, scales from 8B to 32B, and benefits general and vision-dependent reasoning. These results establish native attention as an effective intrinsic signal for perception-aware multimodal RLVR without perturbation-based visual-reliance proxies.

\bibliography{naspo}

\clearpage
\providecommand{\naspo}{NAS-PO}
\providecommand{\nasas}{NAS-AS}
\providecommand{\aac}{AAC}
\providecommand{\old}{\mathrm{old}}
\providecommand{\sg}{\mathrm{sg}}
\providecommand{\eps}{\epsilon}

\appendix

\noindent\textbf{Reading guide.} Appendices~A--B follow the method sequence in the main paper, expanding the attention decomposition, \nasas{}, \aac{}, and their implementation. Appendix~C provides reproducibility details. Appendices~D--F present training dynamics, Visual-Routing Sensitivity (VRS), and qualitative evidence. Appendix~G states the scope and limitations of the claims.

\AppSection{toc-app-a}{Extended Method Derivations}
\label{app:derivations}

This section expands Sections~3.2--3.5 of the main paper in the same order: it first establishes the attention factorization, then details trajectory-level credit shaping, token-level routing consolidation, and adaptive coupling.

\AppSubsection{toc-app-a1}{Causal Native Attention and Lossless Decomposition}
\label{app:decomposition}

For a multimodal prompt $x_i=(I_i,\mathcal{Q}_i)$, let $P_i$ be the prompt length after the visual and textual tokens are serialized, and let $y_i=(y_{i,1},\dots,y_{i,T_i})$ be a sampled response. The log-probability of $y_{i,t}$ is read from query position $p_{i,t}=P_i+t-1$. Writing $\mathcal{C}_{i,t}=\{j:j\le p_{i,t}\}$ for its causal support, the last decoder layer's head-averaged attention is
\begin{equation}
a_{i,t}(j)
=\frac{1}{H}\sum_{h=1}^{H}
\operatorname{Softmax}_{k\in\mathcal{C}_{i,t}}\!\left(
\frac{Q^{(h)}_{i,p_{i,t}}K^{(h)\top}_{i,k}}{\sqrt{d_k}}
\right)_{\!j}.
\end{equation}
Equivalently, the implementation forms the masked logits
\begin{align}
\ell^{(h)}_{i,t,j}
&=\frac{Q^{(h)}_{i,p_{i,t}}K^{(h)\top}_{i,j}}{\sqrt{d_k}}
  +\mathcal{M}_{i,t}(j),\\
\mathcal{M}_{i,t}(j)
&=\begin{cases}
0, & j\le p_{i,t},\\
-\infty, & j>p_{i,t}.
\end{cases}
\end{align}
Numerically, $-\infty$ is replaced by the minimum finite value of the attention-score dtype, and the softmax is evaluated in float32. Thus non-causal positions are removed from the attention support rather than assigned zero logits.

\paragraph{Lossless visual factorization.}
The visual mass collects the attention that falls on the image,
\begin{equation}
    m_{i,t}=\sum_{j\in\mathcal{V}_i} a_{i,t}(j)\in[0,1],
\end{equation}
and defines the visual-mass policy
\begin{equation}
    q_{i,t}=[\,m_{i,t},\,1-m_{i,t}\,].
\end{equation}
When $m_{i,t}>0$ the conditional visual policy is
\begin{equation}
    \rho_{i,t}(j)=\frac{a_{i,t}(j)}{m_{i,t}},\quad j\in\mathcal{V}_i,
\end{equation}
which is a proper distribution over $\mathcal{V}_i$ because $\sum_{j\in\mathcal{V}_i}\rho_{i,t}(j)=m_{i,t}^{-1}\sum_{j\in\mathcal{V}_i}a_{i,t}(j)=1$. The factorization is then exact,
\begin{equation}
    m_{i,t}\,\rho_{i,t}(j)=a_{i,t}(j),\quad j\in\mathcal{V}_i,
\end{equation}
so the pair $(q_{i,t},\rho_{i,t})$ recovers the visual component of the native map without loss whenever the token allocates nonzero visual mass. If $m_{i,t}=0$ then $q_{i,t}=[0,1]$ is still well defined while $\rho_{i,t}$ is undefined and is excluded from conditional alignment. This is precisely why \aac{} aligns $q$ on every valid token but aligns $\rho$ only on visually active tokens.

\paragraph{Why not match the full attention map.}
A full map jointly encodes whether a token consults the image, where it routes visual evidence, and how much non-visual context it reuses. Matching it as a single distribution cannot distinguish image-versus-context allocation from spatial routing. It also imposes a spatial target on tokens with negligible visual mass, for which the conditional distribution is poorly defined. The $(q,\rho)$ split isolates global allocation from spatial routing and lets each factor be supervised only where it is meaningful.

\AppSubsection{toc-app-a2}{NAS-AS: Near-Unit-Mean Advantage Scaling}
\label{app:nasas}

For a rank-local actor mini-batch $\mathcal{B}_r$ with a noncollapsed valid score range, \nasas{} min--max normalizes the visual-reliance scores as
\begin{equation}
    z_i=\frac{s_i-s_{\min}^{(r)}}{s_{\max}^{(r)}-s_{\min}^{(r)}}\in[0,1].
\end{equation}
For an update micro-batch $\mathcal{U}\subseteq\mathcal{B}_r$, let
\begin{equation}
    \bar z_{\mathcal{U}}=\frac{1}{|\mathcal{U}|}\sum_{i\in\mathcal{U}}z_i.
\end{equation}
The upper endpoint is calibrated as
\begin{equation}
    \beta_{\max}^{\mathcal{U}}=\beta_{\min}+\frac{1-\beta_{\min}}{\bar z_{\mathcal{U}}+\eps},
\end{equation}
and each trajectory is scaled by
\begin{equation}
    \alpha_i=\beta_{\min}+z_i\bigl(\beta_{\max}^{\mathcal{U}}-\beta_{\min}\bigr).
\end{equation}
For a nondegenerate update micro-batch with $\bar z_{\mathcal{U}}>0$, the mean scale satisfies
\begin{equation}
\begin{split}
    \frac{1}{|\mathcal{U}|}\sum_{i\in\mathcal{U}}\alpha_i
    &=\beta_{\min}+\bar z_{\mathcal{U}}\bigl(\beta_{\max}^{\mathcal{U}}-\beta_{\min}\bigr)\\
    &=\beta_{\min}+\bar z_{\mathcal{U}}\,\frac{1-\beta_{\min}}{\bar z_{\mathcal{U}}+\eps}\approx 1 .
\end{split}
\end{equation}
The reported setting uses $\beta_{\min}=0.9$. If the rank-local score range collapses, \nasas{} sets $\alpha_i=1$ and leaves the advantages unchanged.

\paragraph{Why the shaped advantage is withheld from \aac{}.}
\nasas{} substitutes $\tilde A_i=\alpha_iA_i$ only into the probability surrogate, while \aac{} gates on the unshaped positive advantage $[A_i]_+$. Sharing $\alpha_i$ with the attention term would let a high-visual-mass trajectory simultaneously receive a larger probability update and a stronger alignment pull, creating a self-reinforcing pressure toward higher attention regardless of task relevance. Withholding $\alpha_i$ from \aac{} avoids applying the same reliance factor twice, while both components still share the visually active indicator and the native-attention pathway.

\paragraph{Why the square-root reliance warp is used.} The excess-mass term applies a square root to the per-token excess, $\sqrt{m^\old_{i,t}-\kappa}$, before averaging. The square root is a concave transform that reduces the influence of isolated high-mass tokens relative to a linear average. Together with the coverage term, it prevents a small number of tokens from dominating the trajectory-level score and favors sustained visual engagement across the response. The same warp reweights tokens when aggregating $\rho$ for the patch ranking in Appendix~\ref{app:vrs}.

\AppSubsection{toc-app-a3}{AAC: Divergence, Gradient Path, and Objective-Space Separation}
\label{app:aac_gradient}

\aac{} aligns the current strategy $(q^\theta,\rho^\theta)$ to the detached reference $(q^\old,\rho^\old)$ on positive-advantage trajectories. The divergence is the Jensen--Shannon divergence
\begin{equation}
    \mathrm{JSD}(p\,\|\,p')=\tfrac{1}{2}\mathrm{KL}(p\,\|\,\bar p)+\tfrac{1}{2}\mathrm{KL}(p'\,\|\,\bar p),
\end{equation}
with $\bar p=\tfrac{1}{2}(p+p')$. For normalized distributions, JSD is symmetric and bounded in $[0,\log 2]$, but this bounded value does not by itself imply bounded derivatives near the probability-simplex boundary. For numerical stability, the implementation clamps each probability coordinate below by $\epsilon_{\mathrm{JSD}}=10^{-10}$ before evaluating logarithms.

The reference-side active support is $c_{i,t}=M_{i,t}\mathbb{I}[m^\old_{i,t}>\kappa]$. Because the current conditional policy also requires nonzero visual mass, the implementation uses
\begin{equation}
    c^\rho_{i,t}=c_{i,t}\,\mathbb{I}[m^\theta_{i,t}>\epsilon_\rho],
    \qquad \epsilon_\rho=10^{-8}.
\end{equation}
The per-token divergence is therefore
\begin{equation}
\begin{split}
    D_{i,t}=\,&\mathrm{JSD}\!\left(q^\theta_{i,t}\,\|\,q^\old_{i,t}\right)\\
    &+c^\rho_{i,t}\,\mathrm{JSD}\!\left(\rho^\theta_{i,t}\,\|\,\rho^\old_{i,t}\right),
\end{split}
\end{equation}
where the reported configuration fixes both branch weights to $\lambda_q=\lambda_\rho=1$ and applies no additional $\rho$ reweighting.

For the reported GRPO configuration, let $\ell_{i,t}=\log\pi_\theta(y_{i,t}\mid x_i,y_{i,<t})-\log\pi_\old(y_{i,t}\mid x_i,y_{i,<t})$. The implementation forms
\begin{equation}
\begin{split}
    r_{i,t}&=\exp\!\left(\mathrm{clip}(\ell_{i,t},-20,20)\right),\\
    r^{\mathrm{clip}}_{i,t}&=\exp\!\left(\mathrm{clip}\!\left(
    \ell_{i,t},\log(1-\varepsilon_{\mathrm{low}}),
    \log(1+\varepsilon_{\mathrm{high}})\right)\right),\\
    w^+_{i,t}&=\min\!\left(r_{i,t},r^{\mathrm{clip}}_{i,t}\right),
\end{split}
\end{equation}
with $\varepsilon_{\mathrm{low}}=0.20$ and $\varepsilon_{\mathrm{high}}=0.28$. For an update micro-batch $\mathcal U$, let $N_{\mathcal U}=\sum_{i\in\mathcal U}\sum_t M_{i,t}$. The reported configuration uses response-token averaging,
\begin{equation}
    L^{+}_{\aac}=\frac{1}{N_{\mathcal U}}
    \sum_{i\in\mathcal U}\sum_t
    M_{i,t}\,[A_i]_+\,\sg(w^+_{i,t})\,D_{i,t}.
\end{equation}
Because $q^\old$, $\rho^\old$, $c^\rho_{i,t}$, $A_i$, and $\sg(w^+_{i,t})$ do not carry gradients, the loss updates only the current attention pathway,
\begin{equation}
    \nabla_\theta L^{+}_{\aac}=\frac{1}{N_{\mathcal U}}
    \sum_{i\in\mathcal U}\sum_t
    M_{i,t}\,[A_i]_+\,\sg(w^+_{i,t})\,\nabla_\theta D_{i,t}.
\end{equation}

\paragraph{Objective-space separation.}
The visual-mass and conditional-routing policies are derived from the same head-averaged attention map and share gradients through the Q/K pathway. Their separation concerns the optimization targets rather than the underlying parameters. The $q$ branch aligns visual-versus-nonvisual allocation on every valid token, whereas the $\rho$ branch aligns within-image routing only on the active visual support. This design provides branch-specific targets and gating without assuming parameter- or gradient-level independence.

\AppSubsection{toc-app-a4}{AAC: Adaptive Coupling and Loss Scale}
\label{app:adaptive_coupling}

The reported implementation computes $\eta$ from detached scalar loss values:
\begin{equation}
    \eta=\mathrm{clip}\!\left(\tau_{\aac}\,
    \frac{|\sg(\mathcal{J}_{\mathrm{PG}}(\tilde A))|}{|\sg(L^{+}_{\aac})|+\eps},\,
    \eta_{\min},\,\eta_{\max}\right),
\end{equation}
where $\tau_{\aac}=0.3$, $\eta_{\min}=0$, $\eta_{\max}=10$, and $\eps=10^{-8}$. If $|L^{+}_{\aac}|\leq 10^{-12}$, the implementation sets $\eta=0$. For nonzero losses and inactive coefficient clipping,
\begin{equation}
    \frac{|\eta L^{+}_{\aac}|}{|\mathcal{J}_{\mathrm{PG}}(\tilde A)|}
    =\tau_{\aac}\frac{|L^{+}_{\aac}|}{|L^{+}_{\aac}|+\eps}
    \approx\tau_{\aac}.
\end{equation}
The rule therefore matches scalar objective magnitudes rather than constraining their gradient-norm ratio. It keeps the routing-alignment term at a controlled scalar scale while leaving verifier-driven policy optimization as the primary objective.

\AppSubsection{toc-app-a5}{Why \naspo{} Avoids Negative Attention Targets}
\label{app:no_negative}

A binary verifier reports only that a final answer is wrong. It does not localize the error to perception, OCR, symbolic manipulation, arithmetic, or formatting. Treating failure as evidence of incorrect visual routing is therefore not justified. \naspo{} lets failed trajectories influence the probability policy through the ordinary negative GRPO advantage, while \aac{} consolidates attention strategies only from positive-advantage trajectories. This matters for vision-language reasoning, where a response may perceive the image correctly and still fail in a later computation step.

\AppSection{toc-app-b}{Training Algorithm, Implementation, and Efficiency}
\label{app:implementation}

\AppSubsection{toc-app-b1}{Full \naspo{} Training Step}
\label{app:algorithm}

\begin{algorithm}[tb]
\caption{NAS-PO: Native Attention-Strategy Policy Optimization}
\label{alg:naspo}
\begin{algorithmic}[1]
\REQUIRE $\pi_\theta$, rollout policy $\pi_{\theta_\old}$, data $\mathcal{D}$, group size $G$, drop quantile $\delta$, minimum scale $\beta_{\min}$, target ratio $\tau_{\aac}$
\STATE Roll out $G$ responses per prompt with $\pi_{\theta_\old}$, verify them, and form advantages $A_i$ \COMMENT{GRPO base}
\STATE Recompute old log-probs and cache detached post-RoPE $Q_\old,K_\old$
\STATE \textbf{Decompose native attention} \COMMENT{lossless $a=m\rho$}
\FOR{each rollout $i$}
    \FOR{each valid token $t$}
        \STATE form $m^\old_{i,t}$, $q^\old_{i,t}$, and $\rho^\old_{i,t}$
    \ENDFOR
\ENDFOR
\STATE Calibrate per-batch floor $\kappa$ and set active mask $c_{i,t}$
\STATE \textbf{NAS-AS: shape advantages} \COMMENT{trajectory level}
\FOR{each rollout $i$}
    \STATE coverage $b_i$, excess mass $u_i$, reliance $s_i=b_iu_i$
\ENDFOR
\STATE Normalize $z_i$, set dynamic $\beta_{\max}$, and form $\alpha_i$, $\tilde A_i=\alpha_iA_i$
\STATE Forward $\pi_\theta$, cache differentiable $Q_\theta,K_\theta$, and rebuild $q^\theta,\rho^\theta$
\STATE Evaluate clipped surrogate $\mathcal{J}_{\mathrm{PG}}(\tilde A)$
\STATE \textbf{AAC: couple attention} \COMMENT{token level, $A_i>0$ only}
\FOR{each rollout $i$ with $A_i>0$}
    \FOR{each valid token $t$}
        \STATE accumulate $D_q$ (all) $+\,c^\rho_{i,t}D_\rho$ (conditional support) into $D_{i,t}$
    \ENDFOR
\ENDFOR
\STATE Form $L^{+}_{\aac}$ with gate $[A_i]_+\,\mathrm{sg}(w^+_{i,t})$ and set $\eta$ by ratio match
\STATE Step on $\mathcal{J}_{\mathrm{PG}}(\tilde A)-\eta L^{+}_{\aac}$ and refresh $\theta_\old\leftarrow\theta$
\end{algorithmic}
\end{algorithm}

Algorithm~\ref{alg:naspo} follows the execution order of one \naspo{} update. It preserves GRPO's rollout, verification, and group-relative advantage computation, then adds three method-specific stages: constructing the detached old attention strategy, shaping trajectory credit with \nasas{}, and reconstructing the differentiable current strategy for \aac{}. Thus the verifier and outer RLVR loop remain unchanged while native attention supplies the trajectory- and token-level routing signals.

\AppSubsection{toc-app-b2}{Post-RoPE Q/K Capture and Response Alignment}
\label{app:qk_capture}

Native attention is reconstructed from the last decoder self-attention layer. \naspo{} captures the query and key tensors after rotary position embedding, since Qwen3-VL applies its query and key normalization and then RoPE before the attention product. Capturing earlier would not reproduce the logits that actually scored each token. Two captures are taken per step:
\begin{itemize}
    \item during old-log-probability recomputation, $(Q_\old,K_\old)$ are detached and stored as behavioral statistics for \nasas{} and as fixed targets for \aac{}.
    \item during the actor update, $(Q_\theta,K_\theta)$ stay differentiable, so the alignment loss backpropagates into the parameters that produce the current attention logits.
\end{itemize}

Response-token alignment follows the autoregressive scoring convention. With full sequence length $S_i=P_i+T_i$, token $y_{i,t}$ is scored by the query at $P_i+t-1$, so the query slice used for response scoring spans positions $P_i-1$ to $S_i-2$. Shifting this slice by one position would attach each attention summary to the wrong generated token. Under grouped-query attention the keys are expanded to the query-head count before averaging. For each sample, a visual-token mask is built from the processor's image and video special tokens. The final-layer attention is computed over the causal support, the visual columns are extracted and head-averaged, and $m$ and $\rho$ are formed. Because different images yield different visual-token counts, the conditional policies are padded to the per-batch maximum. Padding entries carry zero probability and never act as valid targets.

\AppSubsection{toc-app-b3}{Per-Batch Quantile Calibration and Sensitivity}
\label{app:floor}

The reported configuration calibrates $\kappa$ with the \texttt{per\_batch\_quantile} mode. Each data-parallel rank $r$ computes the local $\delta$-quantile of its valid response-token masses,
\begin{equation}
    \kappa^{(r)}=\mathrm{Quantile}_{\delta}
    \big\{m^\old_{i,t}:M_{i,t}=1,\ (i,t)\in r\big\},
\end{equation}
and combines the local estimates using the count-weighted average
\begin{equation}
    \kappa=\frac{\sum_r N^{(r)}\kappa^{(r)}}{\sum_r N^{(r)}},
\end{equation}
where $N^{(r)}$ is the token count on rank $r$. This is a communication-efficient approximation to the pooled global quantile: it requires only two scalar all-reduces (the weighted numerator and the count) rather than gathering every token-level mass. The calibration mask is the response-token mask and $\delta=0.25$. The strict condition $m^\old_{i,t}>\kappa$ defines the active support, so ties at the floor can make the realized active fraction differ slightly from $1-\delta$. Unlike a warmup-then-EMA scheme, this mode recomputes the threshold every step and never freezes it.

\begin{table*}[tb]
\centering
\renewcommand{\arraystretch}{1.2}
{\small\setlength{\tabcolsep}{3pt}%
\begin{tabular}{l|ccccc|c|ccc|c|c}
\toprule
\multirow{2}{*}{\textbf{$\delta$}}
& \multicolumn{6}{c}{\textbf{General Multimodal Reasoning}}
& \multicolumn{4}{c|}{\textbf{Vision-Dependent Multimodal Reasoning}}
& \multirow{2}{*}{\textbf{Overall}} \\
\cmidrule(lr){2-7} \cmidrule(lr){8-11}
& \textbf{MMK12} & \textbf{Geo3k} & \textbf{MathVerse} & \textbf{MathVista} & \textbf{We-Math} & \textbf{AVG}
& \textbf{MMMU-Pro} & \textbf{MathVerse$_V$} & \textbf{LogicVista} & \textbf{AVG} & \\
\midrule
$0$    & 77.11 & 58.52 & 77.23 & 72.26 & 77.71 & 72.57 & 42.64 & 73.70 & 54.25 & 56.86 & 66.68 \\
$0.10$ & 77.06 & 59.23 & \textbf{78.51} & 74.65 & 75.94 & 73.08 & 42.96 & 74.47 & 54.05 & 57.16 & 67.11 \\
\rowcolor{gray!15}
$0.25$ & \textbf{77.49} & \textbf{60.13} & 78.36 & \textbf{75.24} & \textbf{78.09} & \textbf{73.86} & \textbf{43.78} & \textbf{75.61} & \textbf{55.60} & \textbf{58.33} & \textbf{68.04} \\
$0.40$ & 77.25 & 59.02 & 77.73 & 72.20 & 77.39 & 72.72 & 42.45 & 74.42 & 54.58 & 57.15 & 66.88 \\
\bottomrule
\end{tabular}
}
\caption{Sensitivity of GRPO+\naspo{} to the drop quantile $\delta$ on Qwen3-VL-8B (avg@8 acc $\%$). MathVerse$_V$ denotes the vision-centric subset of MathVerse. AVG and Overall are unweighted means over the displayed evaluation configurations. Best results are bolded.}
\label{tab:delta_ablation}
\end{table*}

\paragraph{Sensitivity to the drop quantile.}
Table~\ref{tab:delta_ablation} varies $\delta$ with all other settings fixed. The selected $\delta=0.25$ gives the best general (73.86), vision-dependent (58.33), and overall (68.04) averages. Relative to the $\delta=0$ reference, these improve by 1.30, 1.47, and 1.36 points, respectively. The $0.10$ and $0.40$ settings also outperform $\delta=0$ overall but remain below $0.25$. This trend is consistent with a moderate floor suppressing low-mass noise without discarding useful visual-routing support.

\AppSubsection{toc-app-b4}{Objective Sign Convention and Regularization}
\label{app:loss_sign}

The main text states \naspo{} as a maximization objective,
\begin{equation}
    \mathcal{J}_{\mathrm{NAS\text{-}PO}}(\theta)=\mathcal{J}_{\mathrm{PG}}(\tilde A)-\eta L^{+}_{\aac},
\end{equation}
whereas the implementation minimizes the equivalent loss together with the shared entropy penalty,
\begin{equation}
    L_{\mathrm{NAS\text{-}PO}}(\theta)=L_{\mathrm{PG}}(\tilde A)+\eta L^{+}_{\aac}+\lambda_{\mathrm{ent}}L_{\mathrm{ent}},
\end{equation}
where $\mathcal{J}_{\mathrm{PG}}=-L_{\mathrm{PG}}$ and $L_{\mathrm{ent}}=-\mathbb{E}_{(i,t)}[\log\pi_\theta(y_{i,t}\mid x_i,y_{i,<t})]$ is the sampled entropy penalty. Reference-policy computation is disabled, so no KL term enters the reported objective. The coefficient $\lambda_{\mathrm{ent}}=0.12$ is held identical for GRPO and \naspo{}, isolating native attention-strategy optimization from changes in the shared regularization.

\AppSubsection{toc-app-b5}{Computational Overhead}
\label{app:overhead}

\naspo{} creates no perturbed images, spends no rollout budget on corrupted inputs, and issues no extra reward calls. Relative to plain GRPO it adds only the two post-RoPE captures already listed above and one final-layer attention reconstruction with its JSD evaluation. The reconstruction is confined to the last decoder layer and scales as $O(BHTSd_k)$ for a micro-batch with response length $T$, sequence length $S$, $H$ heads, and head dimension $d_k$, which is far cheaper than perturbation-based methods that need additional image-conditioned forward passes through the full model. The compact 256-token response budget further bounds both the computation and the cached attention-summary size.

\AppSection{toc-app-c}{Experimental Protocol and Hyperparameters}
\label{app:hyperparams}

\AppSubsection{toc-app-c1}{Training and Optimizer Configuration}

Table~\ref{tab:core_config} lists the shared training and evaluation settings of the reported \naspo{}+GRPO 8B and 32B runs. Model-scale differences in hardware and rollout tensor parallelism are shown explicitly. The objective in Eq.~\eqref{eq:nas_po} contains the method-specific policy and attention terms. The entropy penalty in Table~\ref{tab:core_config} is the shared term defined in Appendix~\ref{app:loss_sign} and is applied identically to the corresponding GRPO baselines.

\begin{table}[!t]
\centering
\small
\begin{tabular}{@{}>{\raggedright\arraybackslash}p{0.47\linewidth}>{\raggedright\arraybackslash}p{0.47\linewidth}@{}}
\toprule
\textbf{Category} & \textbf{Setting} \\
\midrule
Base model & Qwen3-VL-8B/32B-Instruct \\
Training data & ViRL39K \\
Reward & Binary answer accuracy \\
Prompt & You first think through the reasoning process. Then, provide your final answer enclosed within \texttt{\textbackslash boxed\{\}} \\
Policy optimizer & GRPO \\ 
Rollouts per prompt & $G=8$ \\
Rollout batch (prompts) & 384 \\
Actor global batch size & 128 \\
Experience micro-batch & 8 per device \\
Update micro-batch & 4 per device \\
Training steps & 202 \\
Random seed & 1 \\
Maximum prompt length & 4096 \\
Maximum response length & 256 \\
Image pixel range & 200,704--1,003,520 \\
Sampling & Temperature $1.0$, top-$p=0.99$ \\
Evaluation & Accuracy@8 \\
Optimizer & AdamW in bf16 \\
Learning rate & $1.0\times10^{-6}$ \\
Weight decay & $1.0\times10^{-2}$ \\
AdamW $(\beta_1,\beta_2)$ & $(0.9,0.999)$ \\
Maximum gradient norm & $1.0$ \\
PPO clipping & Lower $0.20$, upper $0.28$ \\
Reference-policy KL & Disabled \\
Entropy-penalty coefficient & $0.12$ \\
Online group filtering & Group mean accuracy in $(0.01,0.99)$ \\
Rollout tensor parallelism & 1/8 (8B/32B) \\
GPUs & 8/16 NVIDIA H100 (8B/32B) \\
\bottomrule
\end{tabular}
\caption{Training and evaluation configuration for the reported \naspo{} runs.}
\label{tab:core_config}
\end{table}

\AppSubsection{toc-app-c2}{NAS-PO-Specific Hyperparameters}

Table~\ref{tab:naspo_hyperparams} collects the \naspo{}-specific settings. The coverage exponent $\gamma$ generalizes the reliance score to $s_i=b_i^{\gamma}u_i$ and is fixed to $1$, recovering the main-text form $s_i=b_iu_i$. The branch weights $\lambda_q$ and $\lambda_\rho$ are also fixed to $1$, recovering the main-text divergence $D_{i,t}$.

\begin{table}[!t]
\centering
\small
\begin{tabular}{lc}
\toprule
\textbf{Hyperparameter} & \textbf{Value} \\
\midrule
Use \nasas{} & Enabled \\
\nasas{} score source & Native old visual mass $m^\old$ \\
$\beta_{\min}$ & 0.9 \\
Coverage exponent $\gamma$ & 1.0 \\
Excess-mass transform & Square root \\
Use \aac{} & Enabled \\
Attention-token mask & All valid response tokens \\
Attention summary & $q/\rho$ \\
Divergence & JSD \\
$\lambda_q,\lambda_\rho$ & 1.0, 1.0 \\
\aac{} coefficient mode & Adaptive \\
$\tau_{\aac}$ & 0.3 \\
$\eta_{\min},\eta_{\max}$ & 0, 10 \\
Floor mode & Per-batch quantile \\
Drop quantile $\delta$ & 0.25 \\
Floor calibration mask & Response tokens \\
Negative attention margin & Disabled \\
\bottomrule
\end{tabular}
\caption{\naspo{}-specific hyperparameters.}
\label{tab:naspo_hyperparams}
\end{table}

\AppSubsection{toc-app-c3}{Evaluation Protocol}

All reported scores use rule-based exact-match verification rather than LLM-as-a-judge. For each item, we sample eight responses at temperature $1.0$ and top-$p=0.99$ and average their binary exact-match outcomes. We denote this mean response accuracy as Accuracy@8. It is not pass@8. Instances without rule-based verifiable answers are excluded. General, vision-dependent, and overall scores are unweighted means over the displayed evaluation configurations. This protocol preserves the RLVR setting: reward remains sparse and objectively verifiable, while native attention refines credit assignment and visual routing without replacing the verifier.

\AppSubsection{toc-app-c4}{Baseline Provenance and Controlled Comparison}

OneThinker-8B is evaluated using its official Qwen3-VL-8B model. The released NoisyRollout, PAPO, VPPO, and CPPO models use Qwen2.5-VL-7B. To obtain backbone-matched comparisons, we retrain their open-source implementations on Qwen3-VL-8B while retaining the training hyperparameters specified in their released code. All methods are evaluated with the same datasets, prompts, sampling configuration, and exact-match scorers. In the controlled optimizer comparisons, GRPO, GSPO, DAPO, and their \naspo{} variants additionally share the Qwen3-VL policy scale and response-length budget.

\AppSection{toc-app-d}{Training Dynamics and Benchmark-Level Consistency}
\label{app:training_dynamics}

\AppSubsection{toc-app-d1}{Aggregate Accuracy and Reward}

We evaluate GRPO and \naspo{} at 21 checkpoints from step 10 through the final step 202. Each checkpoint is scored with Accuracy@8 on the same eight evaluation configurations used by the main table: five general and three vision-dependent configurations spanning seven benchmarks. Aggregate accuracy is their unweighted average. Training reward is the mean binary verifier reward logged at every optimization step.

\begin{figure*}[t]
\centering
\includegraphics[width=0.98\linewidth]{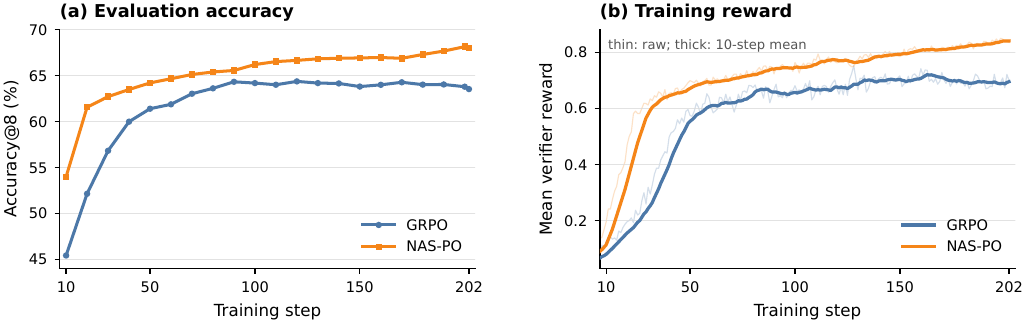}
\caption{Aggregate training dynamics. (a) Unweighted average Accuracy@8 over the eight evaluation configurations. (b) Mean verifier reward at all 202 training steps. Thin lines show raw values, and thick lines show a 10-step moving average. Both single-run trajectories show an early and persistent within-run advantage for NAS-PO over GRPO.}
\label{fig:training_dynamics_appendix}
\end{figure*}

Figure~\ref{fig:training_dynamics_appendix}a shows that \naspo{} exceeds GRPO at all 21 evaluated checkpoints. It reaches GRPO's final Accuracy@8 by step 50, while GRPO first reaches the same level at step 80. Averaged over all checkpoints, \naspo{} obtains 65.32 versus 61.76 for GRPO and finishes at 68.04 versus 63.54 (Table~\ref{tab:training_dynamics_summary}).

The reward trajectory in Fig.~\ref{fig:training_dynamics_appendix}b is consistent with the checkpoint evaluations: \naspo{} is higher at 199 of 202 logged steps, including both the first 30 steps (0.345 versus 0.160 on average) and the final 20 (0.835 versus 0.693). These curves are descriptive single-run trajectories rather than across-seed variance estimates, but their agreement shows that the final gain is not confined to one checkpoint or one metric.

\begin{table}[t]
\centering
\renewcommand{\arraystretch}{1.15}
{\small\setlength{\tabcolsep}{4pt}%
\begin{tabular}{lcc}
\toprule
\textbf{Metric} & \textbf{GRPO} & \textbf{NAS-PO} \\
\midrule
Final average Accuracy@8 & 63.54 & \textbf{68.04} \\
Mean Accuracy@8 over checkpoints & 61.76 & \textbf{65.32} \\
First step reaching 63.54 & 80 & \textbf{50} \\
Mean reward, steps 1--30 & 0.160 & \textbf{0.345} \\
Mean reward, final 20 steps & 0.693 & \textbf{0.835} \\
\bottomrule
\end{tabular}}
\caption{Summary of training dynamics. Accuracy values are percentages.}
\label{tab:training_dynamics_summary}
\end{table}

\AppSubsection{toc-app-d2}{Per-Benchmark Accuracy Curves}

\begin{figure*}[t]
\centering
\includegraphics[width=0.98\linewidth]{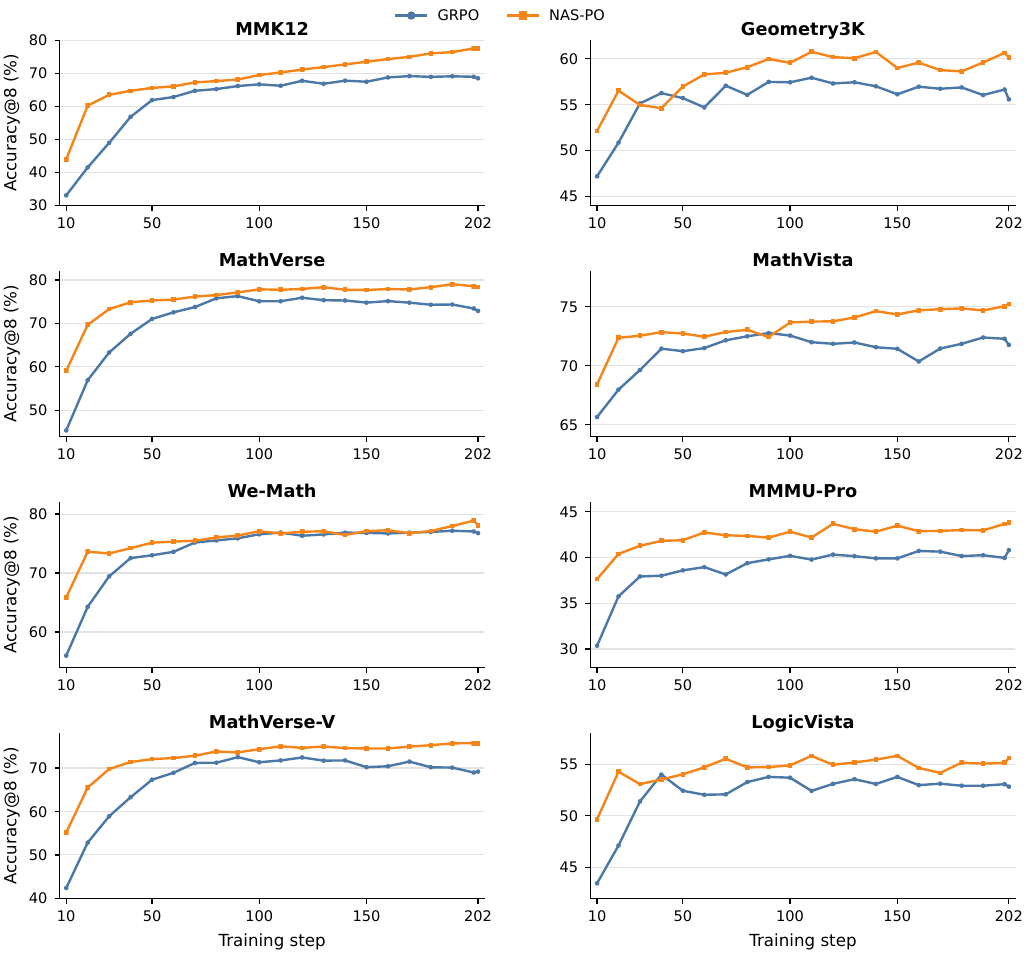}
\caption{Complete checkpoint-wise Accuracy@8 curves. The eight configurations correspond to the seven benchmarks in the main paper, with MathVerse and its vision-dependent subset shown separately. NAS-PO improves the final checkpoint on every configuration and leads on 161 of the 168 configuration--checkpoint pairs.}
\label{fig:per_benchmark_curves}
\end{figure*}

Figure~\ref{fig:per_benchmark_curves} confirms that the aggregate advantage is not driven by a single task. \naspo{} finishes higher on all eight configurations, with final gains of 8.99 points on MMK12, 4.56 on Geometry3K, 5.47 on MathVerse, 3.48 on MathVista, 1.31 on We-Math, 3.00 on MMMU-Pro, 6.43 on MathVerse-V, and 2.77 on LogicVista. Across the full sweep it leads on 161 of 168 configuration--checkpoint pairs, showing that the improvement is broad across mathematical, geometric, logical, and vision-dependent reasoning.

\AppSection{toc-app-e}{Visual-Routing Sensitivity: Protocol and Results}
\label{app:vrs}

Figure~\ref{fig3} tests whether the optimized attention ranking identifies visually load-bearing regions rather than remaining a passive correlate of stronger language priors. The patch-drop probe is applied only at evaluation: it ranks visual patches, removes the most-attended fraction, and measures how often a previously correct answer turns wrong. If the ranking identifies task-relevant routing, removing its top patches should be more disruptive than removing an equal number of bottom-ranked patches.

\AppSubsection{toc-app-e1}{Trajectory-Level Patch Ranking}

For each originally-correct rollout we aggregate the conditional visual policy over answer tokens into a single per-patch importance. Raw $\rho$ contains attention-sink and positional baselines that are not content-specific, so we first subtract a sink estimate. Using a gray-image control that replaces each channel by its mean and preserves the token layout, the teacher-forced sensitivity $v_{i,t}=\log\pi(y_{i,t}\mid I_i)-\log\pi(y_{i,t}\mid I_i^{\mathrm{gray}})$ identifies low-sensitivity filler tokens (bottom $0.4$ quantile). The sink baseline is their mean policy $\rho^{\mathrm{sink}}_i$, and the sink-corrected routing is
\begin{equation}
    \tilde\rho_{i,t}(j)\propto\big[\rho_{i,t}(j)-\rho^{\mathrm{sink}}_i(j)\big]_+,
\end{equation}
renormalized over $\mathcal{V}_i$. Tokens are then weighted by the same reliance warp used in \nasas{}, giving the trajectory aggregate
\begin{equation}
    \bar\rho_i(j)\propto\sum_t c_{i,t}\,\sqrt{\max(m^\old_{i,t}-\kappa,0)}\;\tilde\rho_{i,t}(j),
\end{equation}
again renormalized over $\mathcal{V}_i$. Patches are ranked by $\bar\rho_i$.

\AppSubsection{toc-app-e2}{Perturbation, Re-Rollout, and the VRS Statistic}

Each visual patch is mapped back to its pixel region, and the selected regions are overwritten with the per-image mean color, which removes local content while preserving global color statistics and, critically, the token and position layout so the sequence length is unchanged. For a removed fraction $K$ we build two perturbed images: one dropping the top-$K\%$ most-attended patches and one dropping the bottom-$K\%$ as a control. Re-rolling the same prompt on each (temperature $0.7$, top-$p=0.95$, up to $256$ new tokens, fixed seed) yields a flip indicator that is one when an originally-correct answer becomes wrong. Averaging over originally-correct rollouts gives $\mathrm{flip}_{\mathrm{top}}$ and $\mathrm{flip}_{\mathrm{bot}}$, and the Visual-Routing Sensitivity is their difference,
\begin{equation}
    \mathrm{VRS}=\mathrm{flip}_{\mathrm{top}}-\mathrm{flip}_{\mathrm{bot}}.
\end{equation}
The bottom-$K$ control removes the same number of patches from the same image, so a large VRS isolates the effect of attending to specific regions from the effect of simply deleting image area.

\AppSubsection{toc-app-e3}{Statistical Testing}

For every (method, dataset, $K$) cell we report a percentile bootstrap $95\%$
confidence interval ($10{,}000$ resamples) for $\mathrm{flip}_{\mathrm{top}}$,
$\mathrm{flip}_{\mathrm{bot}}$, and VRS. To compare NAS-PO against each baseline
we pair the two methods on the rollouts both answered correctly, report a paired
bootstrap interval for $\Delta\mathrm{VRS}$, and run a two-sided Wilcoxon
signed-rank test, correcting across the dataset$\times K$ grid with the
Benjamini--Hochberg procedure. Every NAS-PO$-$baseline gap in
Table~\ref{tab:vrs_results} is positive and significant under this correction.
Each cell is capped at $192$ originally-correct rollouts and $K$ is swept over
$\{5,10,15,20,30\}\%$.

\AppSubsection{toc-app-e4}{Results}

Table~\ref{tab:vrs_results} reports VRS at $K{=}15\%$. NAS-PO attains the
highest VRS on every benchmark, averaging $31.43\%$ overall, about $2.3\times$
GRPO's $13.48\%$ and above the strongest perception-aware baselines VPPO
($22.07\%$) and PAPO ($14.62\%$). The margin widens on the vision-dependent split
($38.17\%$ vs.\ $15.75\%$ for GRPO), where correct answers depend most on routing to
the right region. The ordering NAS-PO $>$ VPPO $>$ PAPO $\approx$ GRPO holds at
every swept $K$ (Fig.~\ref{fig3}c), so the result does not hinge on a
particular drop fraction. Because the perturbation is applied only at evaluation,
it tests the consequence of the learned ranking rather than restating the training
signal. The larger top-versus-bottom flip gap shows that \naspo{} ranks visually
load-bearing regions more selectively than the baselines. It supports the relevance
of the learned routing without treating attention as a complete causal explanation
of model computation.

\begin{table*}[t]
\centering
\renewcommand{\arraystretch}{1.2}
{\small\setlength{\tabcolsep}{3pt}%
\begin{tabular}{l|ccccc|c|ccc|c|c}
\toprule
\multirow{2}{*}{\textbf{Method}}
& \multicolumn{6}{c}{\textbf{General Multimodal Reasoning}}
& \multicolumn{4}{c|}{\textbf{Vision-Dependent Multimodal Reasoning}}
& \multirow{2}{*}{\textbf{Overall}} \\
\cmidrule(lr){2-7} \cmidrule(lr){8-11}
& \textbf{MMK12} & \textbf{Geo3k} & \textbf{MathVerse} & \textbf{MathVista} & \textbf{We-Math} & \textbf{AVG}
& \textbf{MMMU-Pro} & \textbf{MathVerse$_V$} & \textbf{LogicVista} & \textbf{AVG} & \\
\midrule
GRPO   & 0.94 & 18.48 & 10.07 & 15.41 & 15.71 & 12.12 & 26.94 & 14.08 & 6.23 & 15.75 & 13.48 \\
PAPO   & 1.00 & 21.59 & 13.14 & 9.48 & 14.38 & 11.92 & 32.37 & 17.61 & 7.41 & 19.13 & 14.62 \\
VPPO   & 5.58 & 41.76 & 18.47 & 18.17 & 16.35 & 20.07 & 39.04 & 27.10 & 10.08 & 25.41 & 22.07 \\
\rowcolor{gray!15}
NAS-PO & 10.64 & 50.06 & 25.11 & 21.48 & 29.61 & 27.38 & 53.16 & 38.73 & 22.62 & 38.17 & 31.43 \\
\bottomrule
\end{tabular}
}
\caption{Visual-Routing Sensitivity at $K{=}15\%$. All entries are percentages, and AVG and Overall are unweighted means over the displayed evaluation configurations.
VRS $=\mathrm{flip}_{\mathrm{top}}-\mathrm{flip}_{\mathrm{bot}}$ on
originally-correct rollouts ($\uparrow$ better). NAS-PO is highest on every
benchmark.}
\label{tab:vrs_results}
\end{table*}

\AppSection{toc-app-f}{Qualitative Analysis}
\label{app:case_study}

The following cases complement the aggregate experiments by illustrating both
successful behavior and residual failure modes. They are qualitative examples
rather than estimates of case frequency.

\AppSubsection{toc-app-f1}{Successful Cases Across Reasoning Domains}

\begin{figure*}[p]
\centering
\includegraphics[width=\linewidth]{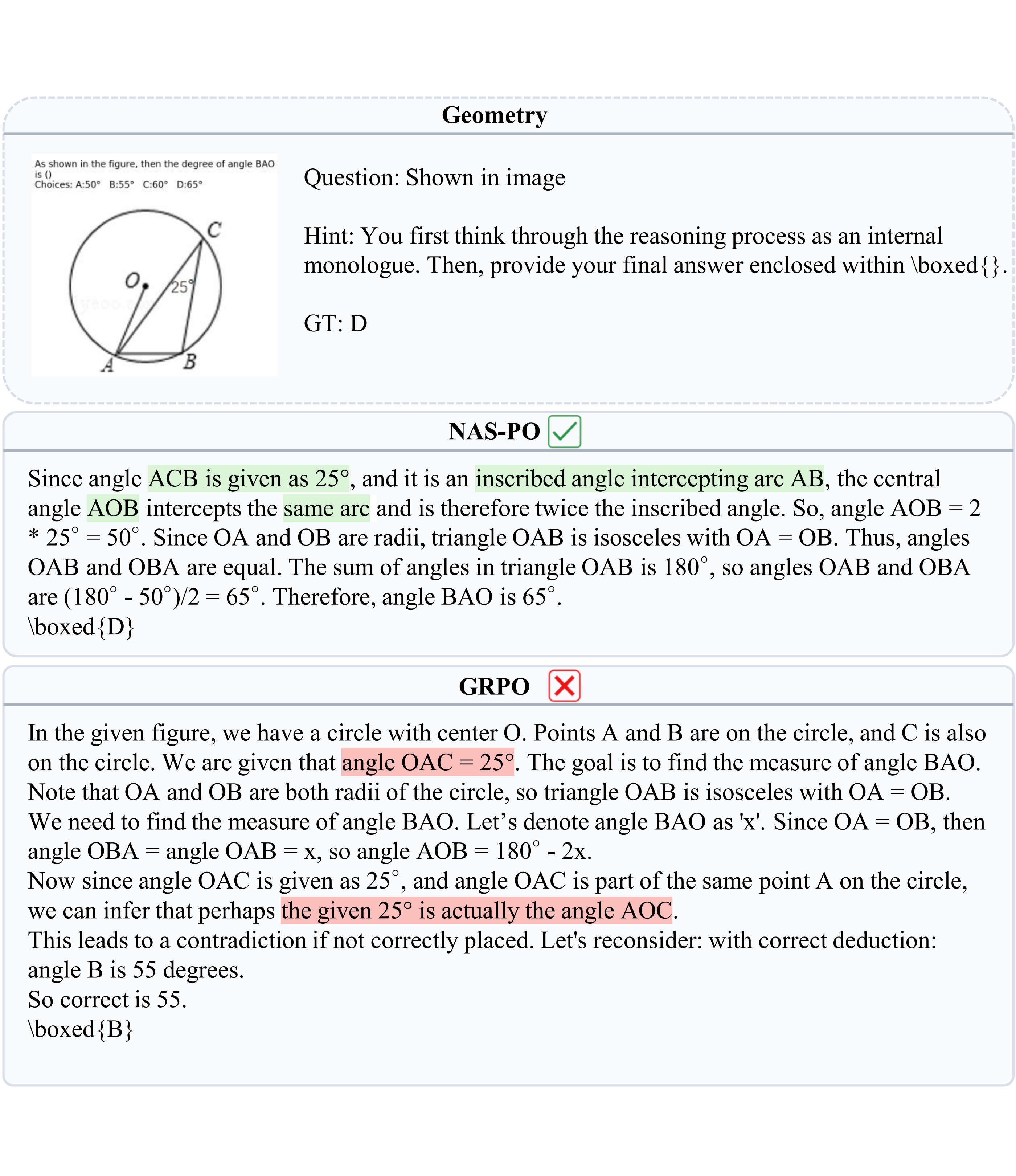}
\caption{Geometry reasoning. NAS-PO correctly treats $\angle ACB=25^\circ$ as an inscribed angle and obtains $\angle BAO=65^\circ$ (D). GRPO misidentifies the marked angle and returns $55^\circ$ (B).}
\label{case1}
\end{figure*}

\begin{figure*}[p]
\centering
\includegraphics[width=\linewidth]{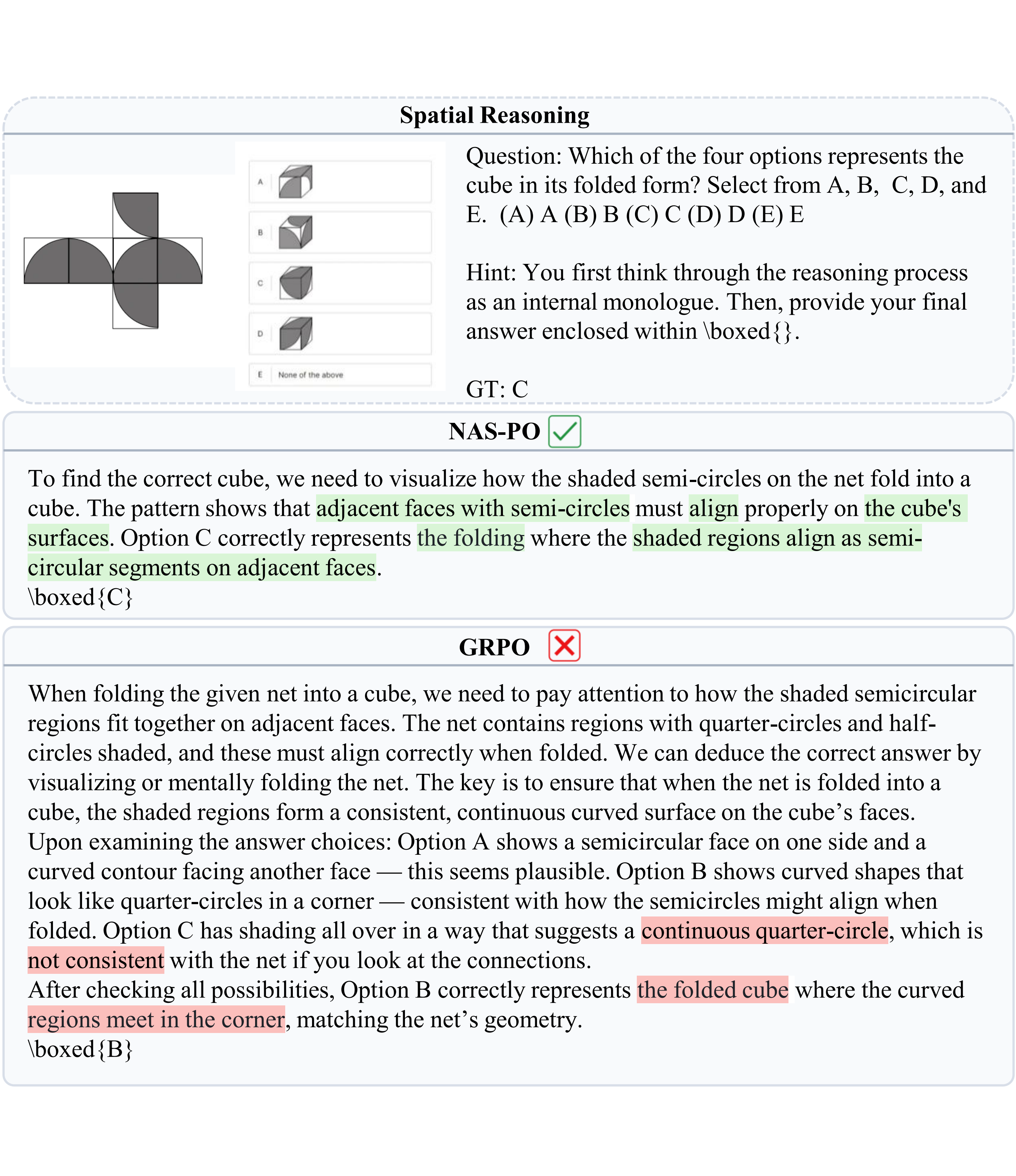}
\caption{Spatial reasoning. NAS-PO correctly tracks how the shaded semicircles meet across adjacent cube faces and selects C. GRPO reconstructs the folded net incorrectly and selects B.}
\label{case2}
\end{figure*}

\begin{figure*}[p]
\centering
\includegraphics[width=\linewidth]{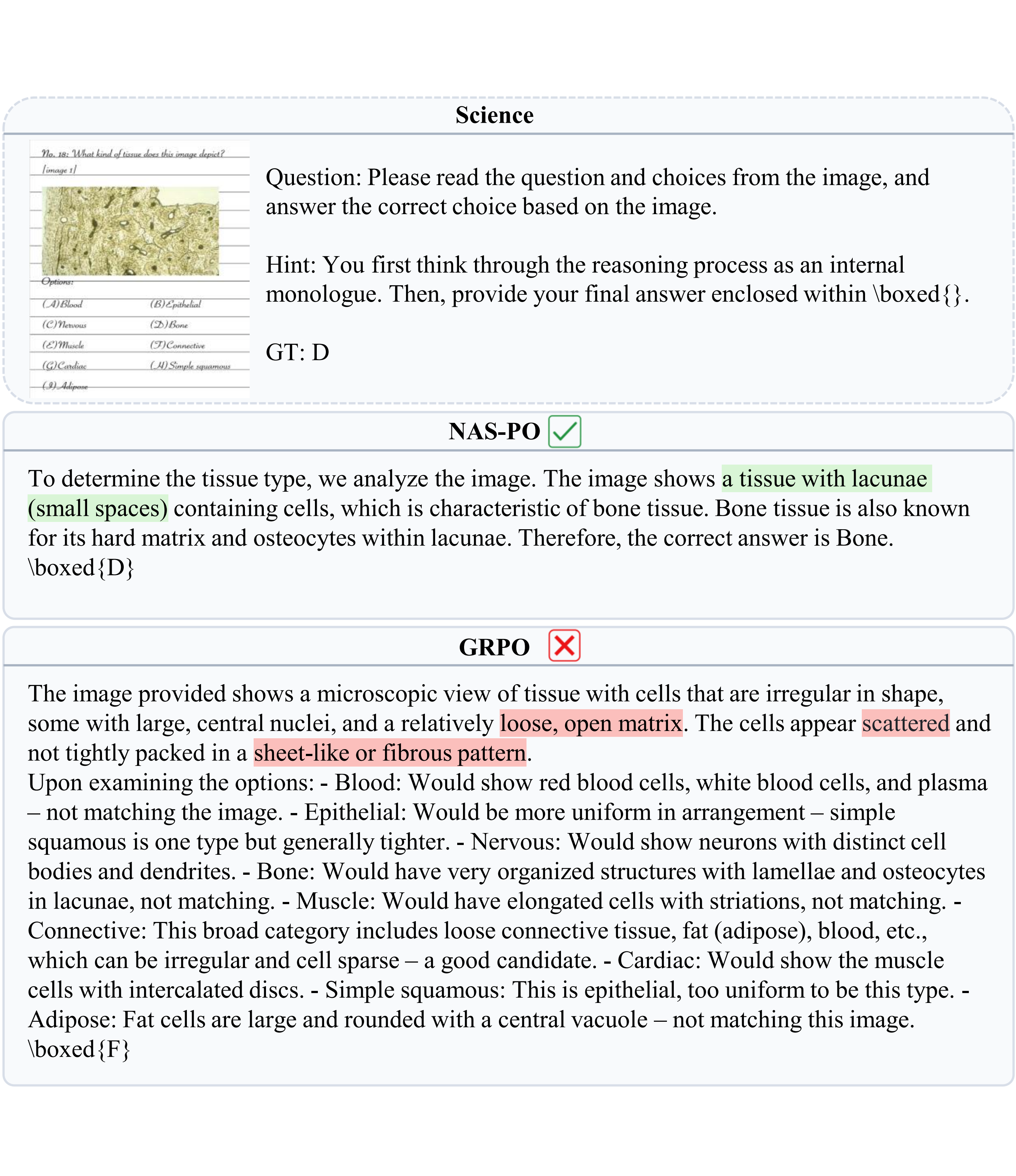}
\caption{Science reasoning. NAS-PO identifies the lacunae characteristic of bone tissue and selects D. GRPO interprets the image as loose connective tissue and selects F.}
\label{case3}
\end{figure*}

\begin{figure*}[p]
\centering
\includegraphics[width=\linewidth]{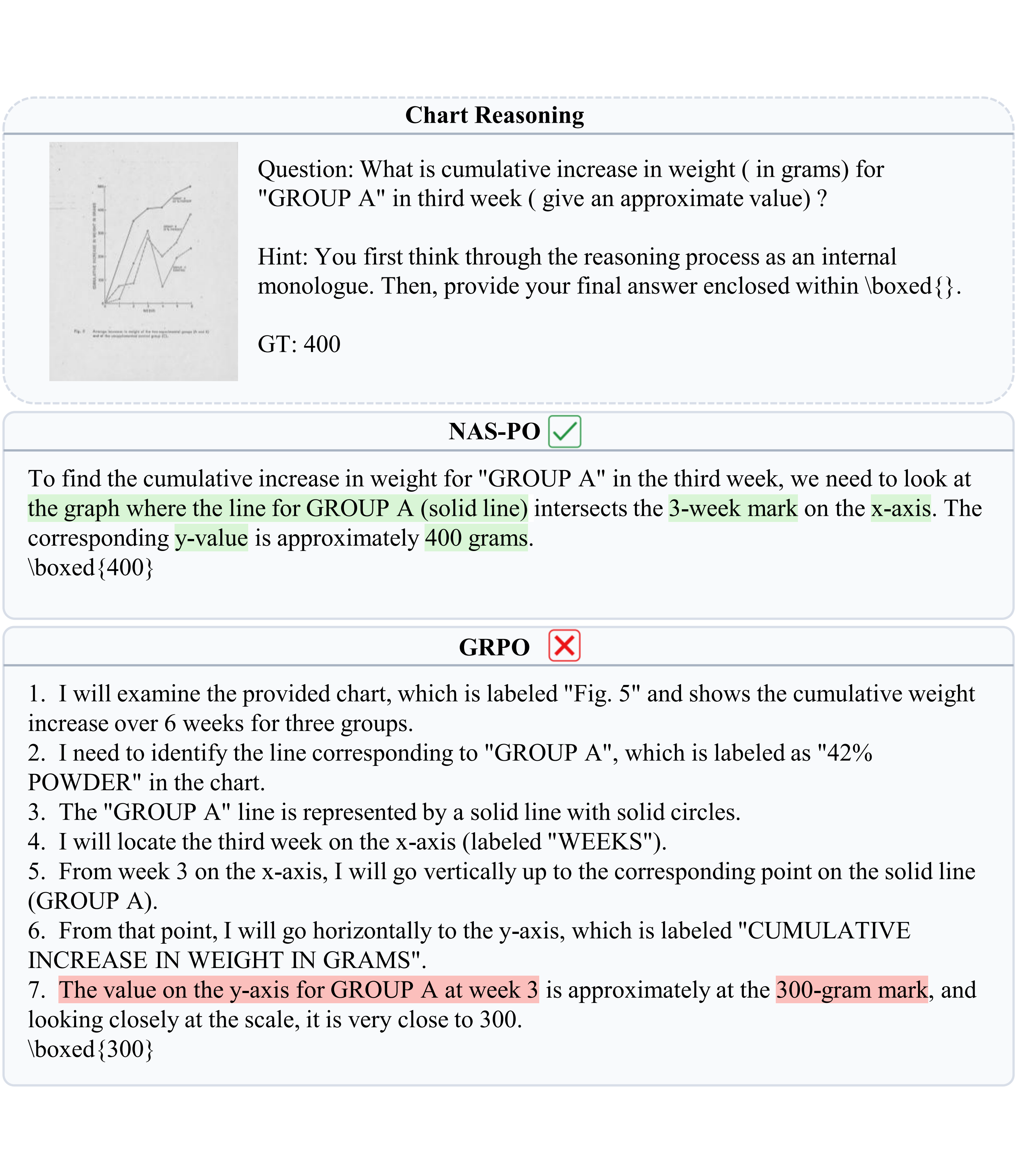}
\caption{Chart reasoning. NAS-PO follows the solid Group A curve to the week-3 value of approximately $400$ grams. GRPO reads the value as approximately $300$ grams.}
\label{case4}
\end{figure*}

Figures~\ref{case1}--\ref{case4} show cases in which
\naspo{} answers correctly while GRPO fails across geometry, spatial, science,
and chart reasoning. The response traces reveal concrete behavioral
differences: GRPO misidentifies the marked angle in Fig.~\ref{case1},
reconstructs the cube net incorrectly in Fig.~\ref{case2}, misclassifies the
tissue structure in Fig.~\ref{case3}, and reads the wrong chart value in
Fig.~\ref{case4}. In each case, \naspo{} recovers the task-relevant visual
relation or value and completes the subsequent reasoning correctly. These
examples complement the aggregate results by showing where the behavioral
gains arise.

\AppSubsection{toc-app-f2}{Residual Perception and Reasoning Failures}

\begin{figure*}[p]
\centering
\includegraphics[width=\linewidth]{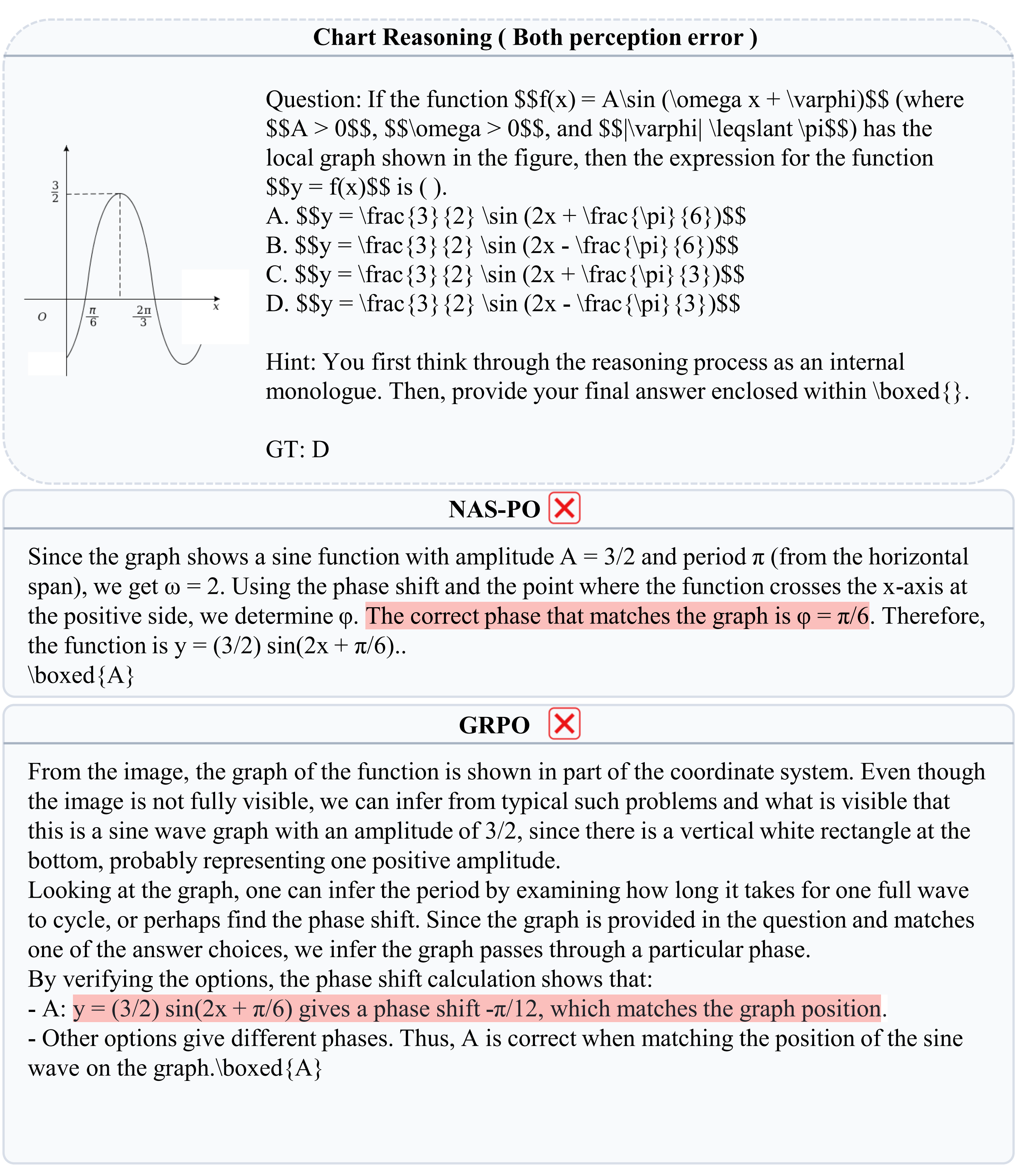}
\caption{Residual perception error. Both NAS-PO and GRPO misinterpret the phase of the plotted sine function and select A instead of the ground-truth option D.}
\label{case5}
\end{figure*}

\begin{figure*}[p]
\centering
\includegraphics[width=\linewidth]{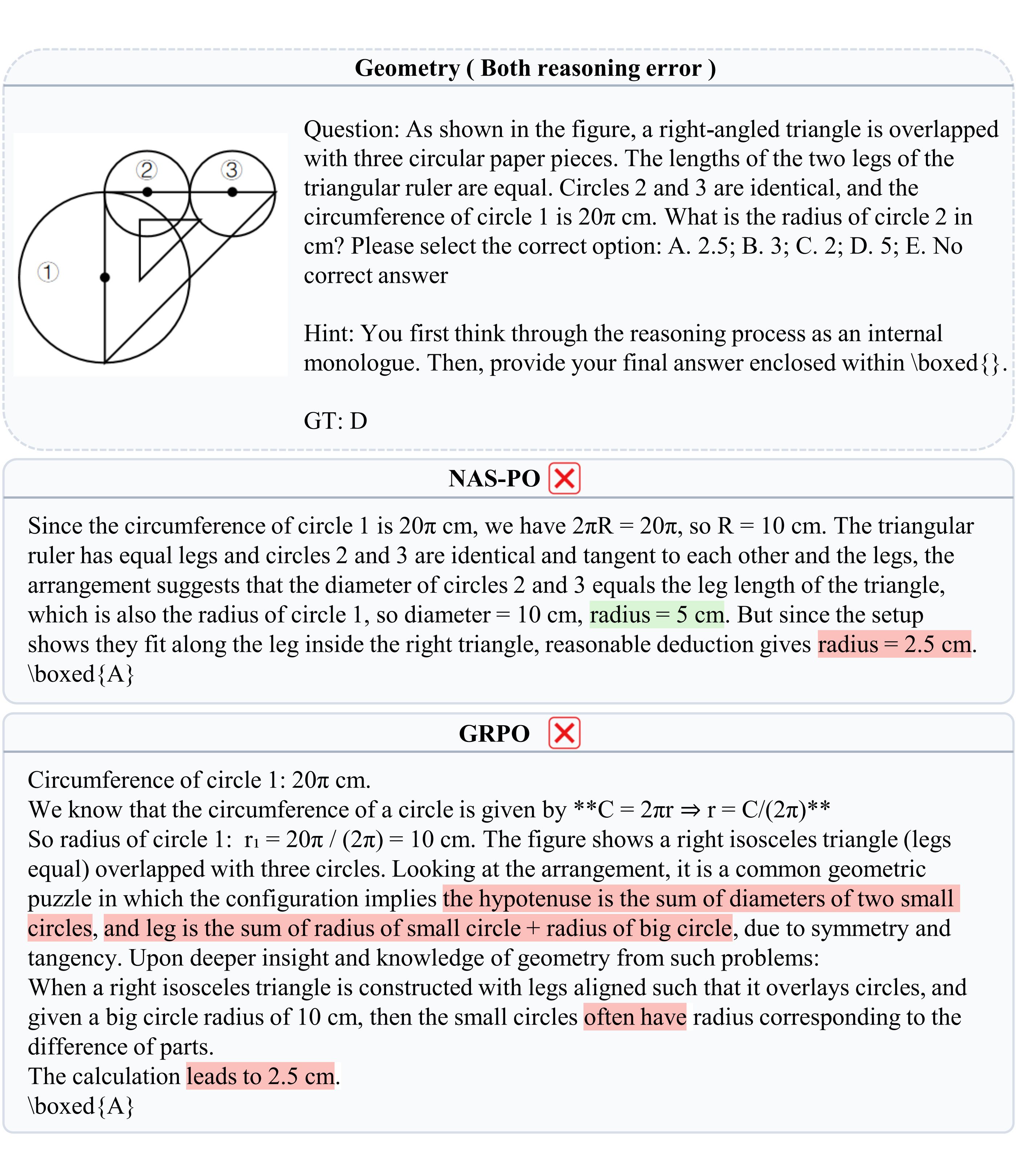}
\caption{Residual reasoning error. NAS-PO first derives the correct radius of $5$ cm but then overrides it and answers $2.5$ cm. GRPO also answers $2.5$ cm. The case separates successful visual extraction from a later reasoning failure.}
\label{case6}
\end{figure*}

Figures~\ref{case5} and~\ref{case6} illustrate two residual failure patterns.
In Fig.~\ref{case5}, both models infer the wrong phase from the sine graph and
select the same incorrect option, suggesting a remaining visual-reading error.
In Fig.~\ref{case6}, \naspo{} first derives the correct radius but then
overrides it without justification, revealing a downstream reasoning
inconsistency even after the necessary visual quantity has been recovered.
This distinction motivates the positive anchoring of \aac{}
(Appendix~\ref{app:no_negative}): a wrong final answer does not reveal whether
the failure arose from visual routing or later reasoning. Failed trajectories
are therefore not treated as negative attention targets.

\AppSection{toc-app-g}{Scope, Limitations, and Future Directions}

\AppSubsection{toc-app-g1}{Interpretation Scope}

\naspo{} does not claim that attention alone fully explains model behavior. The claim is narrower: native attention is the model's own differentiable routing mechanism, and the $(q,\rho)$ decomposition turns it into a structured training signal under sparse RLVR rewards. Training uses only clean rollouts. The perturbations in Appendix~\ref{app:vrs} are evaluation-only interventions that test whether the resulting ranking identifies load-bearing regions, not a complete causal account of model computation.

\AppSubsection{toc-app-g2}{Architecture and Layer Scope}

The method assumes that visual tokens can be identified in the serialized sequence. This holds for the Qwen3-VL interface used here, but architectures with different visual-token conventions require a corresponding visual-support mask. Attention is reconstructed from the final decoder layer, which is efficient and directly aligned with token scoring but may omit earlier-layer perceptual routing. Extending \naspo{} across layers could provide a richer signal at additional memory and computation cost.

\AppSubsection{toc-app-g3}{Feedback Limitations}

\naspo{} is deliberately conservative with failed trajectories. This avoids unreliable negative attention targets, but it also means the attention branch does not learn directly from failures caused by irrelevant visual routing. Richer verifiers or process-level labels could localize such failures and enable reliable negative routing supervision in future work.

\end{document}